\documentclass{article}

\usepackage{microtype}
\usepackage{graphicx}
\usepackage{subfigure}
\usepackage{booktabs} 
\usepackage{cite}

\usepackage{amsthm}
\usepackage{amsfonts}
\usepackage{amsmath}
\usepackage{appendix}
\usepackage{mathtools}
\usepackage{subfigure}
\usepackage{float}
\usepackage[utf8]{inputenc}

\DeclareMathOperator*{\argmax}{arg\,max}
\DeclareMathOperator*{\argmin}{arg\,min}

\DeclarePairedDelimiter{\norm}{\lVert}{\rVert}
\DeclarePairedDelimiter\ceil{\lceil}{\rceil}

\newtheorem{definition}{Definition}

\newtheorem{theorem}{Theorem}

\newtheorem{assumption}{Assumption}

\newtheorem{definition_a}{Definition}
\newtheorem{lemma_a}{Lemma}
\newtheorem{theorem_a}{Theorem}

\newtheorem{remark_a}{Remark}
\newtheorem{fact_a}{Fact}

\newcommand{\sign}{\text{sign}}

\renewcommand{\thesubsection}{\Alph{subsection}}

\usepackage{hyperref}

\usepackage{hyperref}

\usepackage[accepted]{icml2020}

\icmltitlerunning{Towards Deep Learning Models Resistant to Large Perturbations}

\begin{document}
	
	\twocolumn[
	\icmltitle{Towards Deep Learning Models Resistant to Large Perturbations}
	
	
	
	
	\begin{icmlauthorlist}
		\icmlauthor{Amirreza Shaeiri}{sharif}
		\icmlauthor{Rozhin Nobahari}{sharif}
		\icmlauthor{Mohammad Hossein Rohban}{sharif}
	\end{icmlauthorlist}
	
	\icmlaffiliation{sharif}{Department of Computer Engineering, Sharif University of Technology, Tehran, Iran}
	
	\icmlcorrespondingauthor{Amirreza Shaeiri}{shairi@ce.sharif.edu}
	\icmlcorrespondingauthor{Rozhin Nobahari}{nobahari@ce.sharif.edu}
	\icmlcorrespondingauthor{Mohammad Hossein Rohban}{rohban@sharif.edu}
	
	\icmlkeywords{Machine Learning, ICML, Adversarial Training, Large Perturbations, Optimal Robust Classifier}
	
	\vskip 0.3in
	]
	
	\printAffiliationsAndNotice{}
	
	
	
	
	\begin{abstract}
		Adversarial robustness has proven to be a required property of machine learning algorithms. A key and often overlooked aspect of this problem is to try to make the adversarial noise magnitude as large as possible to enhance the benefits of the model robustness. We show that the well-established algorithm called “adversarial training” fails to train a deep neural network given a large, but reasonable, perturbation magnitude. In this paper, we propose a simple yet effective initialization of the network weights that makes learning on higher levels of noise possible. We next evaluate this idea rigorously on MNIST ($\epsilon$ up to $\approx 0.40$) and CIFAR10 ($\epsilon$ up to $\approx 32/255$) datasets assuming the $\ell_{\infty}$ attack model. Additionally, in order to establish the limits of $\epsilon$ in which the learning is feasible, we study the optimal robust classifier assuming full access to the joint data and label distribution. Then, we provide some theoretical results on the adversarial accuracy for a simple multi-dimensional Bernoulli distribution, which yields some insights on the range of feasible perturbations for the MNIST dataset.
	\end{abstract}
	
	\section{Introduction}
	\label{intro}
	
	Modern machine learning models achieve high accuracy on different tasks such as image classification \cite{DBLP:journals/corr/HeZR015} and speech recognition \cite{DBLP:journals/corr/abs-1303-5778, DBLP:journals/corr/XiongDHSSSYZ16a}. However, they are not robust to adversarially chosen small perturbations of their inputs. In particular, one can make imperceptible perturbations of the input data, which can cause state-of-the-art models to misclassify their inputs with high confidence \cite{42503, DBLP:journals/corr/abs-1708-06131, Carlini2016HiddenVC, DBLP:journals/corr/abs-1801-01944}. Also, researchers have shown that even in the physical world scenarios, machine learning models are vulnerable to adversarial examples \cite{DBLP:journals/corr/KurakinGB16, NIPS2019_9362}. The robustness properties of the models in machine learning are a huge concern, as they have been increasingly employed in applications in which safety and security are among the principal issues. Furthermore, the benefits of a robust model go beyond this; for instance, significantly improved model interpretability \cite{tsipras2018robustness} and effective exclusion of brittle features in the learned robust model \cite{ilyas2019adversarial} are some of the model robustness benefits. The former could be observed through an input-dependent saliency map, which is usually a variant of the model gradient with respect to the input, and the fact that this map becomes sparser and semantically more relevant compared to that of a non-robust model. An example of the latter is the ability of the robust model to avoid “shortcut features” that are logically irrelevant but may be strongly correlated to the class. Learning under an adversarial noise would help in obscuring such features and make the model rely on alternative aspects of the input that are more robust for label prediction.
	
	The phenomenon of adversarial machine learning has received significant attention in recent years and several methods have been proposed for training a robust classifier on images. However, most of these methods have been shown to be ineffective \cite{DBLP:journals/corr/CarliniW16, DBLP:journals/corr/abs-1711-08478, DBLP:journals/corr/abs-1802-00420, DBLP:journals/corr/abs-1804-03286, DBLP:journals/corr/abs-1902-02322}, while many others are shown to lack scalability to large networks that are expressive enough to solve problems like ImageNet \cite{DBLP:journals/corr/abs-1902-02918}. To the best of our knowledge, only two training algorithms and their variants, which are called “adversarial training” \cite{Madry2017TowardsDL} and “randomized smoothing” \cite{DBLP:journals/corr/abs-1902-02918}, have been confirmed to be both effective and scalable. Nevertheless, adversarial training remains ineffective for large perturbations of the input \cite{DBLP:conf/iclr/SharmaC18}. More specifically, the authors observed that the adversarial training does not yield an accuracy better than random guessing for perturbation $\ell_\infty$ norm, denoted as epsilon, larger than or equal to $0.4$ in the MNIST dataset. Training in the presence of large perturbations, however, is essential to advance the level of model robustness and its benefits that are gained as a result. For instance, if the model is robust to perturbations of $\ell_0$ norm less than $m$, then only “shortcut features” that are captured in $m$ pixels could be avoided. Otherwise, we would need a larger $m$ to take the best advantage of the model robustness.
	
	
	In addition to the mentioned robust training algorithms, some prior work was also aimed at studying adversarial machine learning from a more theoretical perspective \cite{NIPS2018_7749, DBLP:journals/corr/abs-1906-00555, DBLP:journals/corr/abs-1801-02774, Wang2017AnalyzingTR, ilyas2019adversarial, chen2020data, yin2018rademacher, cullina2018paclearning, zhang2019theoretically, montasser2019vc, diochnos2019lower, attias2018improved, khim2018adversarial}. However, to our understanding, none of them claim to find the optimal robust classifier, assuming knowledge of joint data and label distribution, except only under a strong assumption on the hypothesis space \cite{ilyas2019adversarial}. This helps to find a limit on the perturbation size until we could expect to get a better-than-random classifier assuming sufficiently large training set. 
	
	In section \ref{prop_sec}, we demonstrate that weight initialization of a deep neural network plays an important role in the feasibility of adversarial training of the network under a large perturbation size. A natural question that arises is:
	
	{\it How much can we increase the perturbation $\ell_p$-norm during the training?}
	
	To answer this question, we study the optimal robust classifier, where we have the full knowledge of the input distribution given different classes. In section \ref{theory_sec}, we first prove that, in general, finding the optimal robust classifier in this setting is $\mathcal{NP}$-hard. Therefore, we focus our attention on some conditional distributions, such as the symmetric isotropic Normal and multi-dimensional Bernoulli distributions, in which finding the optimal robust classifier is tractable. Next, we discuss the limits on the perturbation size under which the optimal robust classifier has a better-than-chance adversarial accuracy.
	
	In addition, in section \ref{benef_sec}, following \cite{tsipras2018robustness} and \cite{Kaur2019ArePG}, we show that our models that are trained on larger perturbation sizes have more interpretable saliency maps and attacks, and some other notable visual properties. These results also suggest that using our proposed method to train the model adversarially on a larger perturbation size boosts the benefits that are gained in the robust models. 
	
	\section{Related Work}
	Adversarial machine learning has been studied since two decades ago \cite{ICML-2006-GlobersonR, Dalvi04adversarialclassification, Kolcz2009FeatureWF}. But \cite{Szegedy2013IntriguingPO} and \cite{DBLP:journals/corr/abs-1708-06131} could be considered as the starting point of significant attention to this field, especially in the context of deep learning. Since then, many ideas have emerged that are intended to make the classification robust against adversarial perturbations. However, most of them have later been shown to be ineffective. Gradient masking \cite{papernot2016practical} is an example of issues that arise from such ideas, which leads to a false sense of security. In particular, obfuscated gradients \cite{DBLP:journals/corr/abs-1802-00420} could make it impossible for the gradient-based adversaries to attack the model. Subsequently, these defenses are easily broken by alternative adversaries, such as non-gradient based \cite{Chen_2017, uesato2018adversarial, ilyas2018blackbox}, or black-box attacks \cite{papernot2016practical}. This shows that evaluating the robustness of a neural network is a challenging task. In addition to these unsuccessful attempts, many others are not scalable to large networks that are expressive enough for the classification task on ImageNet and sometimes need to assume specific network architectures \cite{DBLP:journals/corr/abs-1902-02918}.
	
	``Adversarial training” is among the most established methods in the field, which was introduced in \cite{Goodfellow2014ExplainingAH, kurakin2016adversarial} and was completed later by \cite{Madry2017TowardsDL} through the lens of robust optimization \cite{books/degruyter/Ben-TalGN09}. Following this work, there have been numerous studies to improve the adversarial accuracy of adversarial training on the test set by using techniques such as domain adaptation \cite{DBLP:journals/corr/abs-1810-00740} or label smoothing \cite{Shafahi2018LabelSA}, while some others have tried to decrease the computational cost of adversarial training \cite{Wong2020FastIB, DBLP:journals/corr/abs-1904-12843}. Researchers have also tried to apply adversarial training on more natural classes of perturbations such as translations and rotations for images \cite{DBLP:journals/corr/abs-1712-02779}, or on the mixture of perturbations \cite{NIPS2019_8821, maini2019adversarial}.
	Also, \cite{ford2019adversarial} showed the relation between adversarial training and Gaussian data augmentation.

	\section{Proposed Method} \label{prop_sec}
	
	Recall that in adversarial training \cite{Madry2017TowardsDL},
	we need to solve the adversarial empirical risk minimization problem, which is a saddle point problem, that is formulated as:
	\begin{equation}
	\min_{\theta} \sum_{i = 1}^{n} \max_{\delta \in S} \ell(f_\theta(x_i + \delta), y_i)
	\end{equation}
	where $S$ represents the set of feasible adversarial perturbations, and $\ell$ is the loss function. The inner maximization is referred to as the ``adversarial loss”. The setting that we study in this section is when 
	\begin{equation}
	S = \{ \delta \in \mathbb{R}^d : \| \delta \|_{\infty} \leq \epsilon \}, ~~~ \epsilon > 0,
	\end{equation}
	which is the most common perturbation set and used by \cite{Madry2017TowardsDL} and is also considered as the standard benchmark in the context of adversarial examples.
	
	To motivate the proposed method, we would begin with the result of an experiment. We observe that adversarial training on larger epsilons can not decrease the adversarial loss sufficiently, even on the training data. The training adversarial loss for different values of epsilons is provided both on MNIST (Fig. \ref{Fig0}) and CIFAR10 (Supplementary Fig. \ref{app_fig_cifar} in Sec. \ref{appendix_d}).  The detailed experiments are provided in \ref{Exp}. This observation would raise the following question:
	
	{\it Is this an optimization issue in a highly non-convex non-concave min-max game; or is it just that learning is not feasible on large epsilons?}
	
	Before answering this question, one should note that optimization in deep learning has not been extensively studied from a theoretical perspective. Specifically, it is not even completely known why one could achieve near-zero training loss in standard training through randomly initialized gradient descent. This is especially surprising given the highly non-convex loss landscape. Yet the gradient descent is guaranteed to converge in this setting. However, we do not even have this convergence guarantee for adversarial training as a gradient descent-ascent algorithm on the highly non-convex non-concave min-max game \cite{Schfer2019CompetitiveGD}.
	
	Nevertheless, in practice, researchers observed that the trainability of deep models is highly dependent on weight initialization, model architecture, the choice of optimizer, and a variety of other considerations \cite{Li2017VisualizingTL}. The effect of different initializations on adversarial training has not been studied rigorously in the literature \cite{Ben_Daya_Shaifee_Karg_Scharfenderger_Wong_2018, simongabriel2018firstorder}. 
	
	Our main contribution is to propose a novel practical initialization for the adversarial training, which makes learning on larger perturbations feasible. Specifically, for adversarial training on large perturbations, we claim that the final weights of an adversarially trained network on a smaller epsilon can be used for this purpose. Surprisingly, this method can find a good solution for larger perturbations even with a few numbers of training epochs. Fig. \ref{Fig0_1} depicts the above-mentioned idea by illustrating a significant decrease in the training loss for large epsilons using the proposed initialization, which was not possible using a random initialization. This is illustrated in various settings of initial and target epsilons for the MNIST and CIFAR10 datasets. The detailed experiments are provided in \ref{Exp}. 
	

	To gain more insights on the proposed method, we would address these questions: 
	\begin{enumerate}
		\item Does adversarial training converge when we train on larger perturbations? 
		\item Why the low-cost local Nash equilibrium that is found by our method cannot be reached using random initialization?  
		\item Why does the proposed initialization converge quickly with few numbers of training epochs?
	\end{enumerate}
	
	\break
	
	Towards addressing all these questions, in \ref{disc}, we employ the ``loss landscape model”, which gives a geometric intuition about the loss function. 
	
	In \ref{carlini}, we first evaluate our method rigorously and then we discuss another surprising observation, which is the trainability of deep models on very large perturbations (e.g. $\epsilon \geq 0.5$ on MNIST, where the pixel intensities are scaled between $0$ and $1$), which obviously should not be possible, because the attacker can ideally transform all the pixel intensities to a single level and as a result, we should not be able to do better than random guessing. 
		
	In the last section \ref{IAT}, we introduce “iterative adversarial training”, that gradually increases the value of epsilon during the training procedure, as a possible alternative to the proposed weight initialization.

	\subsection{Method Evaluation} \label{Exp}
	
	To demonstrate the effectiveness of our proposed weight initialization on the adversarial trainability of deep models on larger epsilons, we run several experiments on the MNIST \cite{lecun-mnisthandwrittendigit-2010} and CIFAR10 \cite{CIFAR} datasets.
	
	\subsubsection{Experimental setup} \label{Exp_setup}
	
	\begin{itemize}
		\item {\bf Model architecture}: For MNIST, we use a convolutional neural network architecture \cite{44e2afaa580a48bc8b13633b22ff10b4} that is obtained from the TensorFlow tutorial \cite{10.5555/3169991} and for CIFAR10, we use standard ResNet-50 architecture \cite{he2015deep}.
		\item {\bf Base model training}: For training the MNIST models, we scale the pixel intensities between $0$ and $1$. For the optimization, we use the cross-entropy loss, Adam optimizer \cite{kingma2014adam} with learning rate = $0.001$ and batch size = $128$. In standard training, we use only $5$ epochs. For adversarial training, we used $30$ epochs, signed Projected Gradient Descent (PGD) attack \cite{Madry2017TowardsDL} with the random start, PGD learning rate = $0.05$ and number of steps = epsilon$\times 2.5$ / PGD learning rate. For CIFAR10, we scale the pixel intensities between $0$ and $1$. For the optimization, we also use the cross-entropy loss, SGD optimizer with the learning rate schedule being $0.1$ for the first $50$ epochs, $0.01$ for the second $50$, and $0.001$ for the third $50$, all of them with the weight decay = $5 \times 10^{-4}$, batch size = $128$ and also data augmentation. Specifically, for adversarial training, we train for $150$ epochs, based on signed PGD with the random start, learning rate = $2/255$ and steps = epsilon$\times 2.5$ / PGD learning rate. Note that many of these settings are obtained from \cite{robustness}.
		
		\break
		
		\item {\bf Proposed method setup}, which we call as ``Extended adversarial training": For both MNIST and CIFAR10, we trained our models $5$ epochs with the new larger epsilons with the same exact setting as the ones in the base model training, except that for CIFAR10, we do not use data augmentation.
		\enlargethispage{\baselineskip}
		
	\end{itemize}

	
	\begin{figure}
		\includegraphics[width=\linewidth]{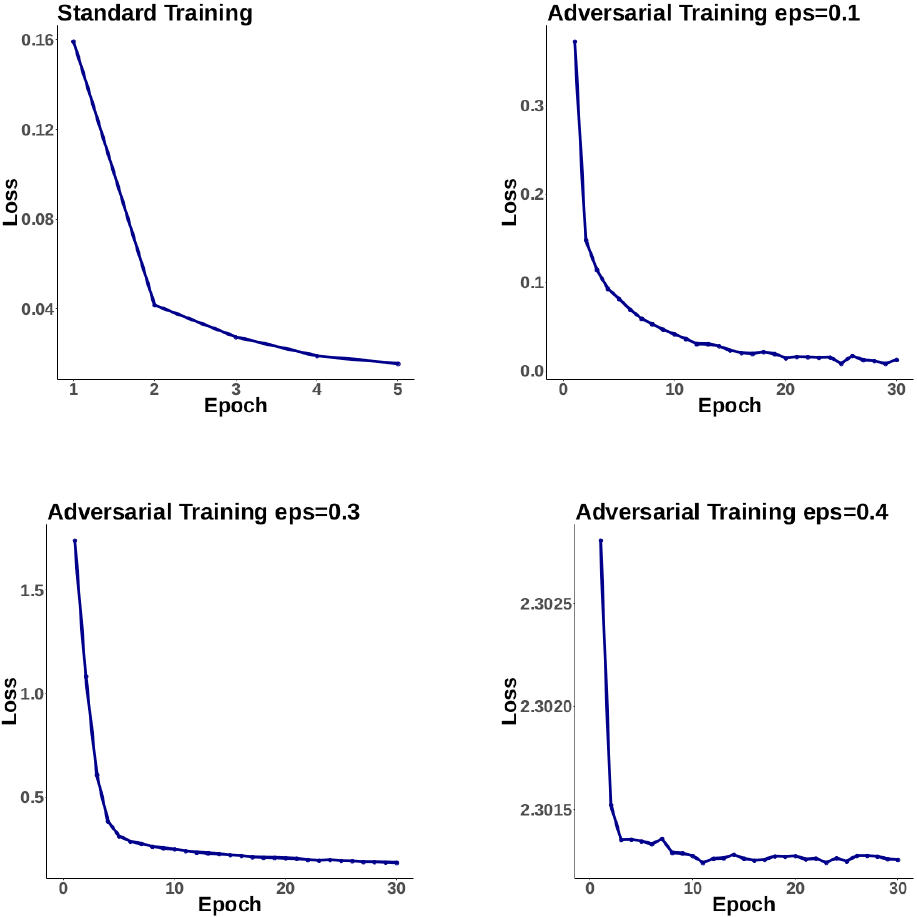}
		\caption{Training adversarial loss against the epoch number for various epsilons in the MNIST dataset. Adversarial training fails to decrease the training adversarial loss on epsilons greater than $\approx 0.35$.}
		\label{Fig0}
	\end{figure}
	
	\begin{figure}
		\includegraphics[width=\linewidth]{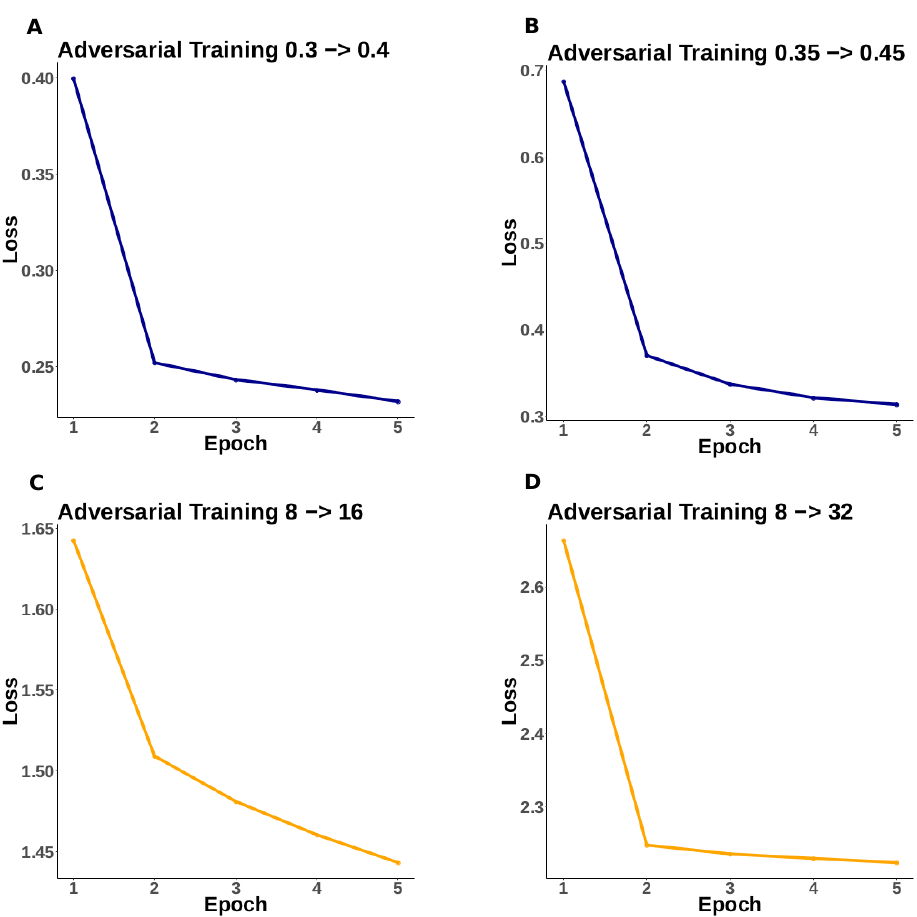}
		\caption{Using the new weight initialization based on an already adversarially trained model on a smaller epsilon makes training on larger epsilons possible. Training from epsilon = $0.3$ to $0.4$ (A), epsilon = $0.35$ to $0.45$ (B), both on MNIST. Training from epsilon = $8/255$ to $16/255$ (C), and epsilon = $8/255$ to $32/255$ (D) both on CIFAR10.}
		\vspace{-1em}
		\label{Fig0_1}
	\end{figure}

	\subsubsection{Experiments}
	
	{\bf Experiment 1}: When we try to apply adversarial training on large epsilons, the training loss does not decrease. We show the training adversarial loss across different epsilons on both MNIST (Fig. \ref{Fig0}) and CIFAR10 (Supplementary Fig. \ref{app_fig_cifar} in Sec. \ref{appendix_d}). We also try to adversarially train the models by adopting different choices of architectures, optimizers, learning rates, weight initialization, and different settings of PGD. These are described in more detail in the Supplementary Materials \ref{appendix_b}.
	
	{\bf Conclusion}: Overall, it seems that these modifications are not playing a major role in the adversarial trainability of deep models on large epsilons. 
	
	{\bf Experiment 2}: We take the weights from an adversarially trained model on epsilon = $0.3$ in MNIST and epsilon = $8/255$ in CIFAR10 as the initial weights, followed by adversarial training on larger epsilons. We report the results in Fig. \ref{Fig0_1}.
	
	{\bf Conclusion}: This initialization makes the adversarial learning possible on larger epsilons in both datasets. 
	
	{\bf Remark 1}: Initializations based on the standard trained models are also observed to be ineffective in this context, which suggests that weights of an adversarially trained model are inherently different from a standard trained model \cite{Goodfellow2014ExplainingAH}. 
	
	\subsubsection{Results}
	
	Now we compare the accuracy of our trained models with standard benchmarks in the literature (see Tables \ref{tab1} and \ref{tab2}). We compared the results to a baseline, namely the model that is trained by the adversarial training with the commonly used epsilons in the literature. Note that one should not expect a model that is trained on a given perturbation size to resist attacks of larger magnitudes. However, this could serve as a simple baseline and give a sense of the adversarial accuracy that is gained through the proposed method. We observe that using our proposed method, we could make adversarial training on larger epsilons feasible and achieve non-trivial and significant adversarial test accuracies. We evaluate all models robustness against the signed PGD attack with $200$ steps, and the other hyper-parameters stay unchanged compared to what we used for training. For the reasons that become clear later in \ref{carlini}, the PGD that is used to solve the inner maximization has a larger step size and smaller learning rate compared to the previously mentioned PGD in \ref{Exp_setup}. It is worth noting that we did not try to fine-tune any of these models to improve the accuracy.
	
	\break

	\begin{figure}[H]
		\includegraphics[width=\linewidth]{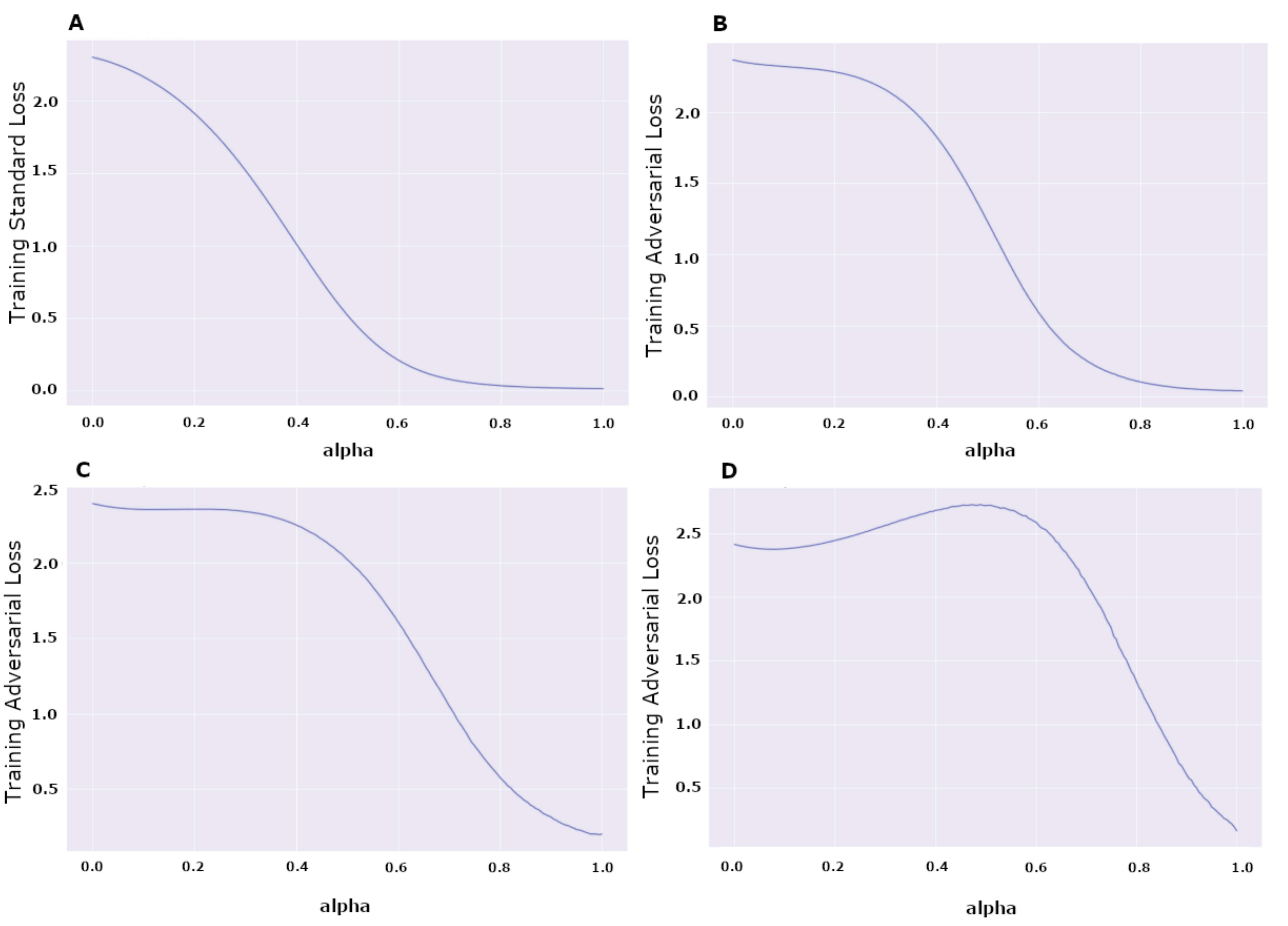}
	    \vspace{-20pt}
		\caption{$\ell_\epsilon(\alpha)$, the interpolated loss function, begins to be non-convex along the direction connecting the initial random to the final trained weights. The plot is made based on 200 points along the horizontal axis. The loss is calculated on a batch of size 512 of the training data. (A) refers to the standard training, while (B), (C), and (D) correspond to epsilon = $0.2$, $0.3$, and $0.4$, respectively.}
		\vspace{-1em}
		\label{Fig1}
	\end{figure}
	
	{\bf Remark}: The adversarial accuracy under the attacks with bounded $\ell_2$ norm are observed to improve in the model that is trained by the proposed idea. More specifically, we consider the evaluation based on a PGD attack with epsilon $= 5$, learning rate $= 0.1$, and $125$ steps. In the MNIST dataset, we obtain an adversarial test accuracy of $\approx 60\%$ for the model that is trained with epsilon $= 0.4$, compared to $30\%$ for the model that is trained with the original adversarial training with epsilon = $0.3$.
	
		\begin{table}[H]
		\caption{Test accuracies of the proposed methods on MNIST compared to that of the Adversarial Training.}
		\label{tab1}
		\vskip 0.15in
		\begin{center}
			\begin{small}
				\begin{sc}
					\resizebox{\columnwidth}{!}{%
						\begin{tabular}{lccccr}
							\toprule
							Model & Clean Accuracy & $\epsilon = 0.3$ & $\epsilon = 0.4$ \\
							\midrule
							$\epsilon = 0.3 \rightarrow 0.4$    & $98\%$ & $94\%$ & $90\%$ \\
							IAT ($\epsilon = 0 \rightarrow 0.4$) (see \ref{IAT}) & $98\%$ & $95\%$ & $91\%$ \\
							Adversarial Training ($\epsilon = 0.3$) & $98\%$ & $91\%$ & $15\%$ \\
							\cite{Madry2017TowardsDL} & & & \\
							\bottomrule
					\end{tabular}}
				\end{sc}
			\end{small}
		\end{center}
		\vskip -0.1in
	\end{table}
	
	\begin{table}[h]
		\caption{Test accuracies of the proposed methods on CIFAR10 compared to that of the Adversarial Training.}
		\label{tab2}
		\vskip 0.15in
		\begin{center}
			\begin{small}
				\begin{sc}
					\resizebox{\columnwidth}{!}{
						\begin{tabular}{lcccccr}
							\toprule
							Model & {Clean Accuracy} & $\epsilon = \frac{8}{255}$ & $\epsilon = \frac{16}{255}$ &  $\epsilon = \frac{32}{255}$ \\
							\midrule
							$\epsilon = {8}/{255} \rightarrow {16}/{255}$    & $75\%$ & $56\%$ & $32\%$ & $10\%$ \\
							$\epsilon = {8}/{255} \rightarrow {32}/{255}$  & $33\%$ & $28\%$ & $24\%$ & $16\%$ \\
							Adversarial Training ($\epsilon = 8/255$) & $87\%$ & $53\%$ & $18\%$ & $1\%$ \\
							\cite{robustness} & & & & \\
							\bottomrule
					\end{tabular}}
				\end{sc}
			\end{small}
		\end{center}
		\vskip -0.1in
	\end{table}
	
	\subsection{Insights on the proposed method} \label{disc}
	
	In this section, we address the three questions that were raised earlier in this section.
	In the deep learning literature, researchers have used loss function visualization to gain insights about the training of deep models, which could not be theoretically explained. Several methods have been proposed for this purpose \cite{Li2017VisualizingTL, 43404}. We further use some visualization to address these questions.
	
	\begin{figure}
		\includegraphics[width=\linewidth]{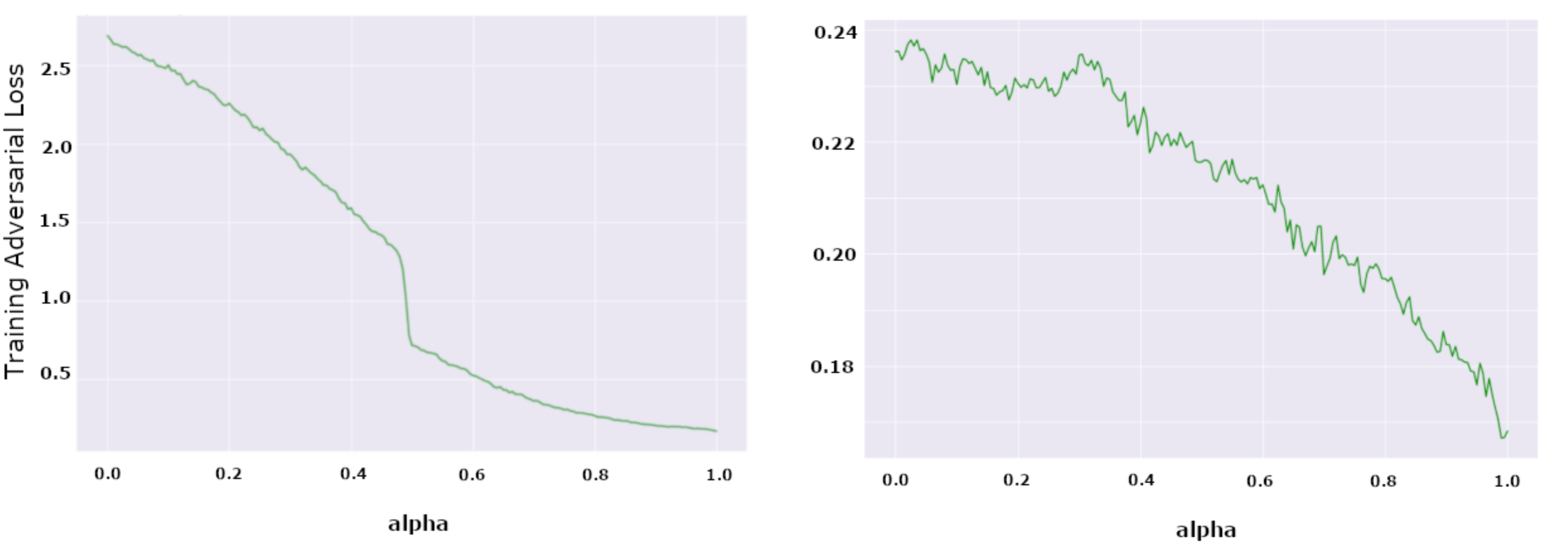}
		\vspace{-20pt}
		\caption{Left: When initialized on weights of an already adversarially trained model with epsilon = $0.3$, $\ell_\epsilon(\alpha)$ could be better approximated by a convex function. Right: $\ell_\epsilon(\alpha)$ based on the initial epsilon stays low as we progress from the initial to the final weights, which implies that the training is focused on low-cost solutions for the initial epsilon (weights that are robust against the initial epsilon) along this line segment.}
		\vspace{-1em}
		\label{Fig2}
		
	\end{figure}

	{\bf Visualization 1}: To assess the convergence of adversarial training, we used two different plots. Specifically, the difference between training adversarial loss, and the $\ell_2$ distance of weights in two consecutive epochs empirically indicate the convergence. The plots are shown in the Supplementary Materials \ref{appendix_e}. 
	
	{\bf Conclusion}: It seems that according to both mentioned plots, the adversarial training empirically has reached a local Nash equilibrium. This result is not unexpected as it has previously been shown that adversarial training would converge irrespective of the epsilon value in a simple model setting \cite{gao2019convergence}.
	
	{\bf Visualization 2}: \cite{43404} proposed a method for loss landscape visualization. Specifically, the adversarial loss function denoted as $\ell_\epsilon(\theta)$, is considered for the weights $\theta$ lying on the line segment that is connecting $\theta_0$ to $\theta_1$, where $\theta_0$ is the initial network weights and $\theta_1$ is the final trained weights. $\ell_\epsilon(\theta)$ could be parametrized by $\alpha$, where $\theta = (1 - \alpha) \theta_0 + \alpha \theta_1$, and $0 \leq \alpha \leq 1$.  
	
	{\bf Conclusion}: 
	\begin{itemize}
		\item We assume that the network initial weights are random. $\ell_\epsilon(\alpha)$ is convex for small perturbations but begins to be non-convex as epsilon grows (Fig. \ref{Fig1}). Therefore, the weights associated with low adversarial loss may not be easily reachable from a random initialization through gradient-based optimization for a large epsilon. 
		
		\item Now, assume that the initial weights are obtained from an adversarially trained model. The $\ell_\epsilon(\alpha)$ seems to be well approximated by a convex function (Fig. \ref{Fig2}; Left). Therefore, solutions with low training adversarial loss become reachable from the initial weights. 
	\end{itemize}
	
	{\bf Visualization 3}: Mode connectivity \cite{Garipov2018LossSM, Drxler2018EssentiallyNB, freeman2016topology} is an unexpected phenomenon in the loss landscape of deep nets. There have been some efforts in the literature to give a theoretical explanation of this effect \cite{DBLP:journals/corr/abs-1906-06247}. We made a similar plot to the visualization 2, with the difference that training adversarial loss is calculated according to the initial epsilon (Fig. \ref{Fig2}; Right). 
	
	{\bf Conclusion}: The given plot suggests that the proposed initialization makes the adversarial training focused on a small weight subspace. Because the training adversarial loss, when evaluated based on the initial epsilon, does not increase, and even decreases slightly. Therefore, we are exploring the weights that are robust against the initial epsilon as opposed to the entire weight space. This suggests ``mode connectivity" in our training and therefore, we could expect the optimization to converge quickly.
	
	\subsection{Evaluating adversarial robustness} \label{carlini}
	
	As already mentioned, evaluating against the adversarial attacks has proven to be extremely tricky. Here, we are inspired by the latest recommendations on evaluating adversarial robustness \cite{DBLP:journals/corr/abs-1902-06705}. Note that our threat model is simple, and is indeed similar to the one that is used in the evaluation of adversarial training. Motivated by the recently recommended checklist for evaluation of adversarial robustness, we applied various sanity checks such as black-box attacks \cite{papernot2016practical}, gradient-free \cite{Chen_2017}, brute force attacks, and a novel semi-black-box attack to make the threat model broader. We found out that the proposed defenses pass all these checklists on MNIST. The details of the evaluation setup and the results for these experiments are explained in the Supplementary Materials \ref{appendix_c}. 
	
	Building upon the mentioned recommendation list, we further evaluated our models based on more challenging tests. These are designed to increase our confidence in the claims that we make about the model robustness that is achieved by the proposed method. 
	
	We first decrease the learning rate of PGD to $0.01$ and increased the number of steps to $1000$ to find a stronger attack. In the model that is trained based on our proposed initialization with target epsilon = $0.5$, this led to adversarial accuracy of $30\%$, as opposed to $80\%$ that is obtained when the same PGD as in the training is used for the test-time attack. However, the mentioned stronger attack does not significantly affect the adversarial accuracy of the model with target epsilon = $0.4$. To make the attack even stronger we used the actual gradient, as opposed to the signed gradient in PGD \cite{Madry2017TowardsDL}, with the learning rate = $5 \times 10^4$ and step number $= 2000$. We observed that the adversarial accuracy dropped to $14\%$ - $20\%$ for the former model, but this again did not affect adversarial accuracy of the model with target epsilon = $0.4$. Notably, we could not decrease validation adversarial accuracy of the model with the target epsilon = $0.4$ although we tried out various PGD settings.
	
	\begin{figure}
		\includegraphics[width=\linewidth]{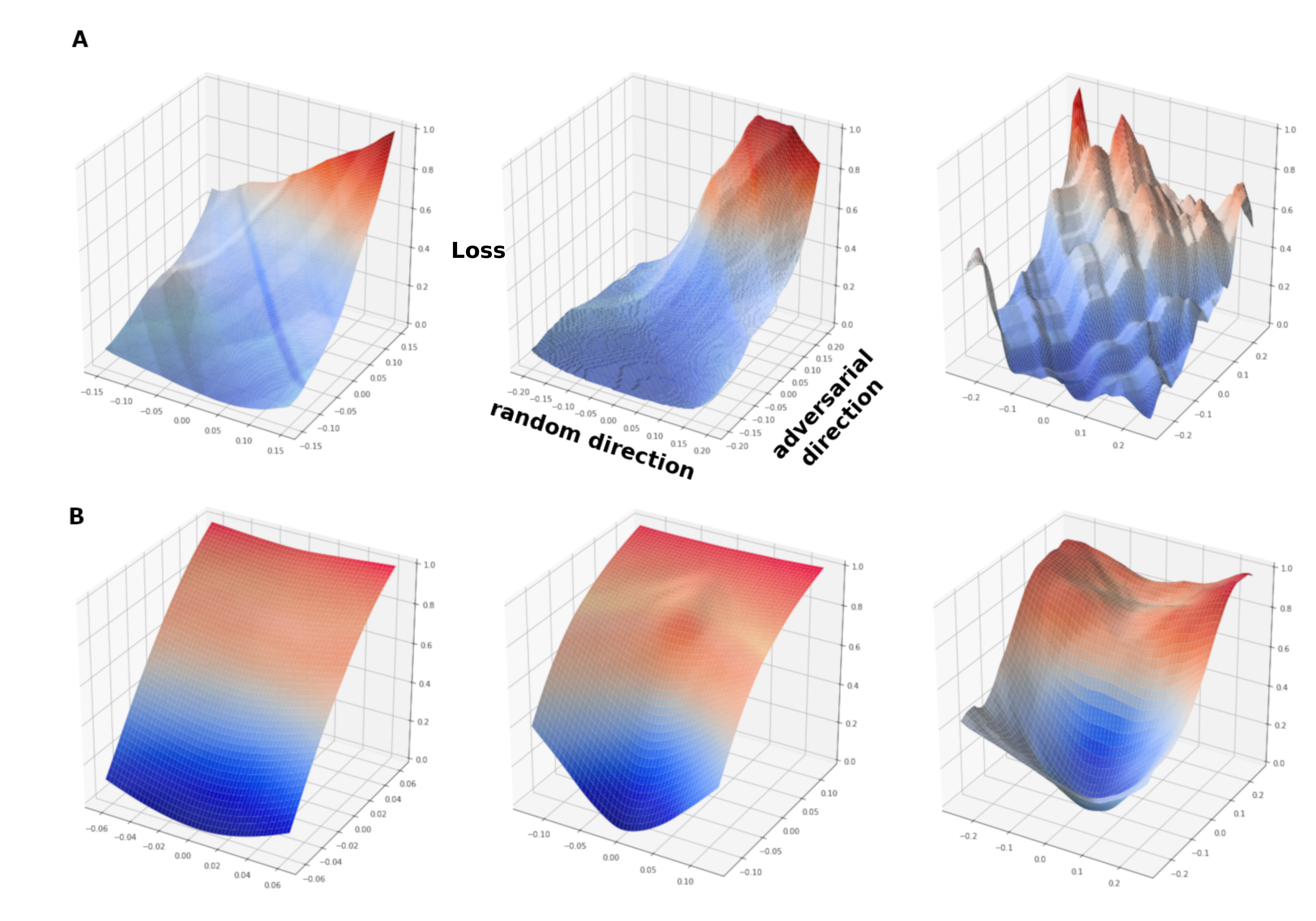}
		\caption{The loss function is visualized along the random and adversarial directions on a test data point. (A): The plots for the MNIST data, for epsilon = $0.3$,  $0.4$, and $0.5$ from left to right. The models do not show significant signs of gradient obfuscation, which is a sudden jump or non-smoothness of the loss, for epsilon up to around $0.4$. (B): In CIFAR10, the plots are made for models that are trained with epsilons $16/255$, $32/255$, $64/255$ from left to right. We do not observe gradient obfuscation up until $64/255$. For MNIST, we used $10^4$ points on the horizontal plane and we considered half of the corresponding epsilon along both directions. For CIFAR10, we used $50^2$ points and considered epsilon along both directions. Note that, {\it in the MNIST dataset}, the ``clean" loss landscape is not observed on all test samples even in the original adversarial training. However, for the majority of such samples, we get the desired loss landscape.}
		\label{Fig3}
	\end{figure}

	One could use the loss function visualization along the adversarial and random direction in the input space to assess the gradient obfuscation. This plot suggests that the model with the target epsilon = $0.4$ does not exhibit gradient obfuscation, while the model with epsilon = $0.5$ shows signs of gradient obfuscation (Fig. \ref{Fig3}). 
	
	To go further, we used the PGD with learning rate = $0.005$ and $200$ steps for the inner maximization in adversarial training from epsilon = $0.3$ to target epsilon = $0.5$. We noted that the adversarial training loss increased in the final model from $0.6$ to $1.2$, and test adversarial accuracy based on the same PGD decreased significantly from $80\%$ to $30\%$. Surprisingly, unlike the model that is trained with a weaker PGD, we could successfully attack the trained model with the stronger PGD even with the weaker attack that was introduced in \ref{Exp_setup}. 
	
		\begin{figure*}
		\begin{center}
			\includegraphics[width=0.8\linewidth,]{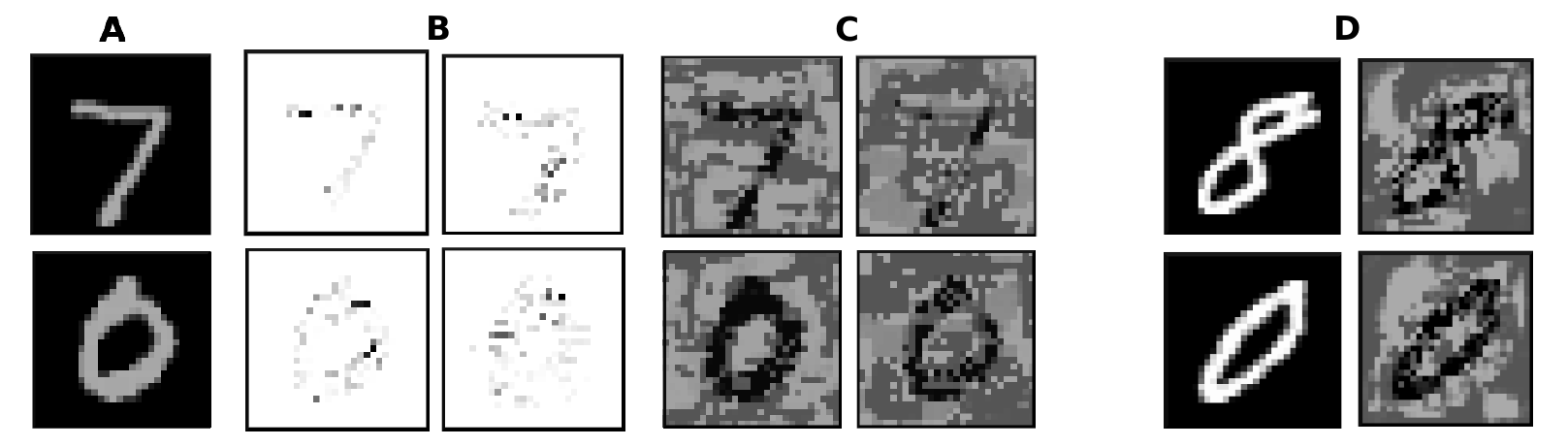}
		\end{center}
		\caption{(A): Two test samples from MNIST. (B) The loss gradient (saliency map) with respect to the perturbed input (of epsilon = 0.3) that is obtained for the model trained on epsilon = $0.3$ (Right) and the model that is trained on epsilon from $0.3$ to $0.4$ (Left). (C) The perturbation (of size epsilon = $0.45$) for the corresponding models in (B). (D) Left: the original image. Right: the perturbations that are obtained from the model trained on epsilon from $0.3$ to $0.5$, based on the attack obtained from the actual gradient that is mentioned in Sec. \ref{carlini}. We note that the attack that is based on the actual gradient exhibits better interpretability even in the model trained with the original adversarial training. Recall that this model shows obvious signs of gradient obfuscation. Note that we did not apply random restart in any of these plots, and the images are scaled for visualization purposes. Overall, training on larger perturbations leads to stronger interpretability of both the model and the attacks.}
		\label{Fig5}
	\end{figure*}

	\break

	This is analogous to the remedy that is used to avoid gradient obfuscation in FGSM by increasing the number of steps, which led to emergence of the PGD attack. Indeed, the first attempt to train the model using a PGD with a small number of steps has overfitted to this weak attack. We believe that by making the inner maximization more accurate, one could not even be able to train a model at epsilon = $0.5$. 
	
	Note that for CIFAR10, as the adversarial accuracy is already low for large epsilons, we do not need to evaluate the model on the stronger attack. We included the gradient obfuscation plots for CIFAR10 in Fig. \ref{Fig3}, which shows a similar trend to those of MNIST.
	
	\subsection{Iterative adversarial training (IAT)} \label{IAT}
	
	Inspired by previously suggested ``extended model training”, we propose a new method, which we call ``Iterative Adversarial Training” (IAT). In IAT, the epsilon is gradually increased from 0 to the target epsilon with a specific schedule across epochs during the training.
	
	\begin{figure}[H]
		\begin{center}
			\includegraphics[width=0.45\linewidth, ]{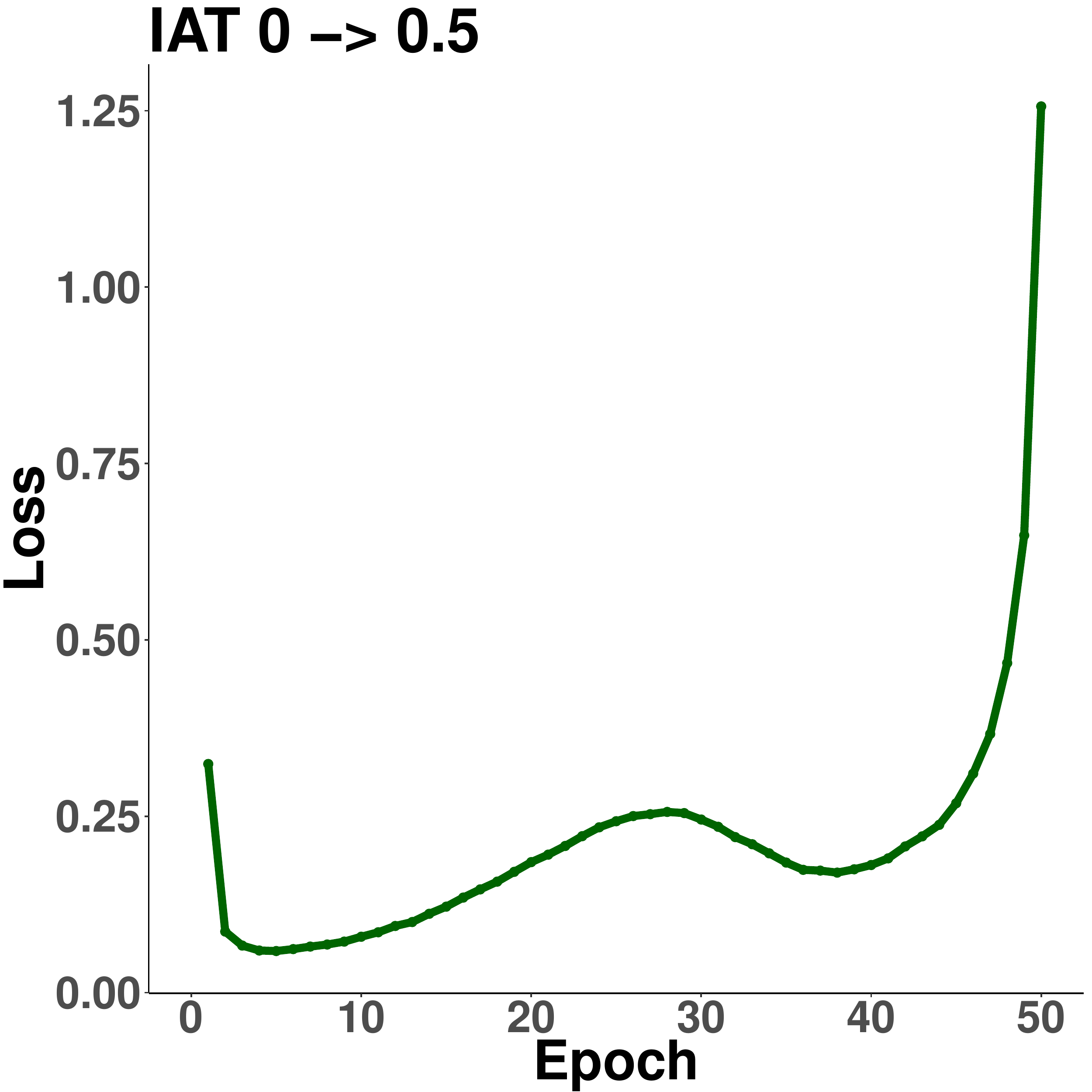}
		\end{center}
		\caption{Training adversarial loss against the number of epochs for the proposed iterative adversarial training (IAT) algorithm on MNIST.}
		\label{Fig4}
	\end{figure}
	
	For the MNIST dataset, we used the same model as described earlier. The only difference here is that we increase epsilon by $0.01$ in each epoch of adversarial training. In addition, we could identify the stopping point for epsilon by plotting the training adversarial loss against epochs and stop as soon as a significant sudden increase is observed (Fig. \ref{Fig4}).
	
	For the CIFAR10 dataset, we tested several schedules (e.g. linear and exponential schedules) with different settings, but unfortunately, the trained models either did not show a good adversarial accuracy or got overfitted.
	
	\section{Benefits of Adversarial Robustness} \label{benef_sec}
	
	The interpretability of the machine learning models has emerged as an essential property of the model in many real-world applications. These, include areas where an explanation on the model prediction is required for either validation or providing overlooked insights about the input-output association to a human user. In addition to the recent efforts to understand the internal logic of these models \cite{DBLP:journals/corr/SelvarajuDVCPB16, DBLP:journals/corr/MontavonBBSM15}, it has been observed that interpretability comes as an unexpected benefit of adversarial robustness. For instance, the saliency map that highlights influential portions of the input becomes sparser and more meaningful in the robust models \cite{tsipras2018robustness}. 
	
	The last layer of a deep neural network can be thought of as the representation that is learned by the model. This representation could be divided into a set of robust and non-robust features \cite{ilyas2019adversarial}. Robust features are the ones that are highly correlated to the class label even in the presence of adversarial perturbations. These features, therefore, have a close relation to the human perception and changes in these features can affect the meaning of data for a human being. In contrast, non-robust features show a brittle correlation to the class label and therefore are not aligned with the human perception. During the training process, a network has a bias towards learning non-robust features \cite{ilyas2019adversarial, engstrom2019a}. Adversarial training is a natural remedy that prevents the network from learning and relying on non-robust features. By training a network on lager perturbations, we narrow the set of non-robust features that classifier can learn and as a result, the loss gradient with respect to the input relies on more robust features and become more interpretable. Therefore, we would expect more relevant saliency maps for such a model. 
	
	Next, we assess this property in our learned models and compare them to models that are robust to smaller perturbations (Fig \ref{Fig5}). Specifically, we observe that the saliency map that is obtained from the loss gradient with respect to the input, evaluated on the perturbed data, in the model trained on larger epsilon, becomes more compact and concentrated around the foreground. Relevant to this, we also plotted the adversarial perturbation on the model that is trained on the MNIST dataset with epsilon = $0.4$ and compared it to a network that is trained on epsilon = $0.3$. The PGD attack is based on epsilon = $0.45$. We observe that the attack on both models has aimed to obscure the digit, but the attack from the model with higher epsilon would erase the digit more precisely. It has been reported that the attack on adversarially trained models lacks interpretability \cite{schott2018adversarially, DBLP:conf/iclr/SharmaC18}, however, our model appears to yield more interpretable attacks compared to the baseline. 
	
	\section{Theoretical Results} \label{theory_sec}
	
	We first briefly recap the definition of classification error rate and classification adversarial error rate and then, we define 1. optimal classifier, 2. $(p, \epsilon)-$optimal robust classifier, 3. optimal classification error rate and 4. $(p, \epsilon)-$optimal classification error rate with respect to the given distribution. Next, we discuss such optimal classifiers and their corresponding error rates under two specific distributions.
	
	\subsection{Basics and definitions}
	
	In the subsequent definitions, we let $L = \{ 1, 2, \ldots , c \}$ be the set of labels. We also, let $D : \mathbb{R}^d \times L \rightarrow \mathbb{R}$ denote the joint feature and label distribution.
	
	\begin{definition}
		The error rate of a classifier $h : \mathbb{R}^d \rightarrow L$ on the distribution $D$ is defined as:
		\begin{equation}
		R(h, D) := \mathbb{E}_{x, y \ \sim \ D} \left\{ \mathbb{I}(h(x) \neq y) \right\},
		\end{equation}
		where $\mathbb{I}(.)$ is the indicator function that takes $1$ if its input is a true logical statement, and $0$ otherwise.
	\end{definition}
	
	\break
	
	\begin{definition}
		The classifier $h : \mathbb{R}^d \rightarrow L$
		is said to be an optimal classifier on $D$ if for any other classifier $h^{\prime} : \mathbb{R}^d \rightarrow L$, 
		$R(h, D) \leq R(h^{\prime}, D)$.
		$R(h, D)$ is also called the optimal Bayes classification error rate on $D$.
	\end{definition}
	
	\begin{definition}
		Let $x \in \mathbb{R}^d$ be a point in the input space.
		Let $p, \epsilon \in \mathbb{R}^{+}$.
		Then, the perturbation set $B_p^{\epsilon}(x)$ is defined as:
		\begin{equation}
		B_p^{\epsilon}(x) := \left\{ x^{\prime} \in \mathbb{R}^d \; : \; 
		\norm{x - x^{\prime}}_{p} \leq \epsilon \right\},
		\end{equation}
		where $\norm{.}_{p}$ is the $\ell_p$ norm of a vector. 
	\end{definition}

	\begin{definition}
		Let $B_p^{\epsilon}(x)$ be a perturbation set.
		Then, adversarial error rate of a classifier $h : \mathbb{R}^d \rightarrow L$ on the perturbation set $B_p^{\epsilon}(x)$ and the distribution $D$ is defined as:
		\begin{equation}
		R_{p, \epsilon}^{(adv)}(h, D) := \mathbb{E}_{x, y \ \sim \ D} \left\{ \mathbb{I}(\exists \delta \in B_{p}^{\epsilon} : h(x + \delta) \neq y) \right\}.
		\end{equation}
	\end{definition}

	\begin{definition} \label{rob_opt}
		The classifier $h : \mathbb{R}^d \rightarrow L$ is said to be a $(p, \epsilon)$-optimal robust classifier on $D$ if for any other classifier $h^{\prime} : \mathbb{R}^d \rightarrow L$, $R^{(adv)}_{p, \epsilon}(h, D) \leq R^{(adv)}_{p, \epsilon}(h^{\prime}, D)$. $R^{(adv)}_{p, \epsilon}(h, D)$ is also called the $(p, \epsilon)$-optimal classification error rate on $D$.
	\end{definition}

	As the optimal and the optimal robust classifiers tend to solve completely different problems by the definition, they can be different functions for a given distribution. One can easily show that they can be different functions in the general case. Consider a very simple distribution on $X \subset \mathbb{R}^d$ and $L = \{1, -1\}$, which contains just two points $x_1$ and $x_2$, $D(x_1, 1) = D(x_2, -1) = 1/2$. Assume that $\| x_1 - x_2 \|_p \leq \epsilon$. The optimal classifier outputs $1$ and $-1$, when the input is $x_1$ and $x_2$, respectively. However, the $(p, \epsilon)$-optimal robust classifier outputs a constant, either $1$ or $-1$. 
	
	\subsection{Optimal classifiers}
	
	In the standard setting, we can find the optimal classifier for a given distribution by assuming full access to the joint data and label distribution efficiently, specifically, through applying the Bayes optimal classifier theorem \cite{book1997}. However, the problem of finding the $(p, \epsilon)-$optimal robust classifier in this setting is computationally hard in general.
	
	\begin{theorem}
		The problem of finding the optimal robust classifier given the joint data and label distribution $D(x, y)$ for the perturbations as large as $\epsilon$ in $\ell_p$ norm, is an $\mathcal{NP}$-hard problem.
	\end{theorem}
	
	It was shown that adversarial robustness might come at the cost of higher training time \cite{Madry2017TowardsDL}, requiring more data \cite{NIPS2018_7749}, and losing the standard accuracy \cite{tsipras2018robustness}. In addition, we show that finding the optimal robust classifier is an $\mathcal{NP}$-hard problem. Note that this computational difficulty is different from the ones that arise in finding the global minimum of a non-convex function. Because by assuming an infinite amount of training data, instead of minimizing the non-convex empirical loss, we can well approximate the data distribution and therefore, we can efficiently find an optimal Bayes classifier. However, in this case, finding an optimal robust classifier is still computational prohibitive in the general case.
	
	We note that for several distributions, one could efficiently find the optimal robust classifier. These include the Gaussian and mixture of Bernoulli distributions.
	
	\subsection{The Isotropic Gaussian model}
	
	\begin{definition}
		Isotropic Gaussian class conditional distributions: We first sample $y \in \{ -1, 1\}$ uniformly and then sample a data point in $\mathbb{R}^d$ from a multivariate Gaussian distribution $N(\mu_{y}, \sigma^2 I)$, where without loss of generality $\mu_1 = -\mu_{-1}$.
	\end{definition}
	
	\begin{theorem}
Let $(x | y) \sim N(\mu_y, \sigma^2 I)$, and $y \in \{-1, 1\}$ with equal probabilities, then assuming that the $(2, \epsilon)$-optimal robust classifier has a continuous decision boundary, $f(x) = 2 \mathbb{I}(w^{\top} x \geq 0) - 1$ is $(2, \epsilon)$-robust , where $w = \mu_1 - \mu_{-1}$. Furthermore, the optimal Bayes and optimal robust classification error rates are $\Phi\left(\frac{-\| \mu_1 - \mu_{-1} \|_2}{2\sigma} \right)$, and $\Phi\left(\frac{-\| \mu_1 - \mu_{-1} \|_2 + 2\epsilon}{2\sigma} \right)$, respectively, where $\Phi$ is the cumulative distribution function for the standard normal distribution.
	\end{theorem}
	
	\subsection{The multi-dimensional Bernoulli model}
	
	\begin{definition}
		Uniform mixture of two multivariate Bernoulli distributions, $\text{Bernoulli}(\theta^{(1)}, \theta^{(-1)}, t)$: Let $\theta_i^{(1)}$ and $\theta_i^{(-1)} \in \{ -1, 1 \}$ for all $1 \leq i \leq d$, and $t \in \mathbb{R}^{+}$. We first randomly sample the class label $y$ uniformly from $\{-1, 1 \}$. Then, we sample each dimension of $x$ according to a Bernoulli distribution:
		\begin{equation}
		x_i = 
		\begin{cases}
		\theta_i^{(y)}, & \text{ with probability } 1/2 + t \\
		-\theta_i^{(y)}, & \text{ with probability } 1/2 - t 
		\end{cases}
		\end{equation}
		Therefore, the conditional distribution of $x$ given $y$ becomes:
		\begin{equation}
		D(x | y) = (1/2 + t)^{d - \| x - \theta^{(y)} \|_0} (1/2 - t)^{\| x - \theta^{(y)} \|_0}
		\end{equation}
	\end{definition}
	
	\begin{theorem} \label{lemma_bayes_main}
		Let $\text{Bernoulli} \ (\theta^{(1)}, \theta^{(-1)}, t)$ be a Bernoulli model.
		Let $(x, y) \sim \text{Bernoulli}(\theta^{(1)}, \theta^{(-1)}, t)$. Then, 1. the optimal Bayes' classifier $h^\star : \{-1, 1\}^d \rightarrow \{-1, 1\}$ on this distribution is of the form:	
		\begin{equation}
		h^\star(x) = 
		\begin{cases}
		1, & \| x - \theta^{(1)} \|_0 \leq \| x - \theta^{(-1)} \|_0 \\
		-1, & \text{otherwise},
		\end{cases}
		\end{equation}
		and 2. if $k$ is even, the optimal classification error rate is:	
		\begin{multline}
		~~ \sum_{i = 0}^{k/2} {k \choose i} (1/2 + t)^{i} (1/2 - t)^{k - i} \\
		~~~ - 1/2 {k \choose k/2} (1/2 + t)^{k/2} (1/2 - t)^{k/2}
		\end{multline}
		Otherwise, if $k$ is odd, the classification optimal error rate would be:
		\begin{equation}
		\sum_{i = 0}^{\lfloor k/2 \rfloor} {k \choose i} (1/2 + t)^{i} (1/2 - t)^{k - i},
		\end{equation}	
		where $k := \| \theta^{(1)} - \theta^{(-1)} \|_0$, which is the number of dimensions that $\theta^{(1)}$ and $\theta^{(-1)}$ do not agree. We call these dimensions as the ``effective dimensions". 
	\end{theorem}
	
	\begin{assumption} \label{symm_asm}
		We assume that for $ \forall u, v \in \{ -1, 1\}^d$ and $\| u^{\prime} - \theta^{\prime (1)} \|_0 = \| v^{\prime} - \theta^{\prime (1)} \|_0$, the $(0, \epsilon)$-optimal robust classifier would give the same label to $u$ and $v$. Note that, this is a rational assumption because of the symmetry of the problem.
	\end{assumption}

	\begin{theorem}
		Let $\text{Bernoulli} \ (\theta^{(1)}, \theta^{(-1)}, t)$ be a Bernoulli model.
		Let $(x, y) \sim \text{Bernoulli}(\theta^{(1)}, \theta^{(-1)}, t)$.
		Let $B_0^{\epsilon}(x)$ be the perturbation set.
		Then, under the \hbox{assumption \ref{symm_asm},} 1. the $(0, \epsilon)$-optimal robust classifier $h^\star : \{-1, 1\}^d \rightarrow \{-1, 1\}$ on this distribution is of the form:
		\begin{equation}
		h^{\star (adv)}(x) = 
		\begin{cases}
		1, & \| x^\prime - \theta^{\prime (1)} \|_0 \leq s \\
		-1, & \text{otherwise},
		\end{cases}
		\end{equation}
		where $s \in \{0, 1, ..., k\}$ is a threshold, and $x^\prime$, and $\theta^{\prime (1)}$ are the restrictions of $x$, and $\theta^{(1)}$ to the effective dimensions, defined in theorem \ref{lemma_bayes_main}, respectively, 
		and 2. the classification $(0, \epsilon)$-optimal error rate would be:
		\begin{multline}
		R_{0, \epsilon}^{(adv)}(h, D) =
		1/2 \sum_{i = 0}^{s + \epsilon} {k \choose i} (1/2 + t)^{i} (1/2 - t)^{k - i} + \\ 1/2 \sum_{i = 0}^{k - s + \epsilon - 1} {k \choose i} (1/2 + t)^{i} (1/2 - t)^{k - i}
		\end{multline}
	\end{theorem}
		
	\begin{theorem}
		Let $\text{Bernoulli} \ (\theta^{(1)}, \theta^{(-1)}, t)$ be a Bernoulli model.
		Let $(x, y) \sim \text{Bernoulli}(\theta^{(1)}, \theta^{(-1)}, t)$.
		Let $B_\infty^{\epsilon}(x)$ be a perturbation set.
		Then, 1. the $(\infty, \epsilon)$-optimal robust classifier $h^{\star (adv)} : \{-1, 1\}^d \rightarrow \{-1, 1\}$ on this distribution for $\epsilon < 1$ is of the form: 
		\begin{equation}
		h^{\star (adv)}(x) = 
		\begin{cases}
		1, & \| T(x) - \theta^{(1)} \|_0 \leq \| T(x) - \theta^{(-1)} \|_0 \\
		-1, & \text{otherwise},
		\end{cases}
		\end{equation}
		where $T(x) = (\sign(x_1), \ldots, \sign(x_d))$ and $\sign(r)$ is $1$ if $r \geq 0$ and is $-1$ otherwise,
		and 2. the classification $(\infty, \epsilon)$-optimal error rate would be:
		\begin{equation}
		R_{\infty, \epsilon}^{(adv)}(h, D) = 
		\begin{cases}
		R(h^{\star}, D), & \epsilon < 1 \\
		1/2, & \text{otherwise},
		\end{cases}
		\end{equation}
		where $h^{\star}$ is the optimal Baye's classification rule.
	\end{theorem}
	{\bf Remark 1}: Note that the effective dimension, $k$, rather than the real dimension of the input, appears in the equations for the optimal classifiers.
	
	{\bf Remark 2}: For the case of isotropic Gaussian distribution, for the $\ell_2$ attack model, maximum allowed adversarial perturbation, before getting the trivial random chance adversarial accuracy is half of the $\ell_2$ distance between the centers of the two Gaussians. For the mixture of multi-dimensional Bernoulli distribution, however, for the $\ell_\infty$ attack model, such a limit is $\epsilon = 1$, which translates to $\epsilon = 0.5$ for the mixture elements being in $\{0, 1\}$. For the $\ell_0$ attack, one has to use the final equation of the proof to find the mentioned limit for epsilon, which is plotted in the Supplementary Materials \ref{appendix_g}.
	\\
	\\
	Detailed proofs of the theorems in this section are provided in Supplementary Materials \ref{Appendix_a}.
	
	It is notable that, there is some work in the literature \cite{gourdeau2019hardness,bubeck2018adversarial,mahloujifar2018adversarially, garg2019adversarially}, which discussed the computational hardness of adversarial robust learning. However, none of them considered the optimal case.
	
	\section{Conclusion and future work}
	
	We demonstrated that the weight initialization plays an important role in the adversarial learnability of deep models on larger perturbations. Specifically, weights from an already robust model on a smaller perturbation size could be an effective initialization to achieve this goal. In spite of the promises in the proposed idea, some directions remain unstudied. We next provide a list of possible future directions to extend this work:
	\begin{itemize}
		\item In this work, we demonstrated the importance of weight initialization in the feasibility of adversarial training on large perturbations. However, more theoretical explanation is still essential to better understand this phenomenon from an optimization perspective, e.g. explaining the geometry of loss landscape.
		\item We suggest that the idea behind this method may be the key to make learning possible on large variance Gaussian noise data augmentation. This can be used to improve methods such as the randomized smoothing \cite{DBLP:journals/corr/abs-1902-02918, salman2019provably} as well.
		\item We showed that finding the optimal robust classifier is an $\mathcal{NP}$-hard problem in general. We also showed that it can be computed for some specific distributions. However, it would be useful to understand what general properties of the data distribution make finding the optimal robust classifier efficiently possible.
		
	\end{itemize}
	
	\section*{Code}
	You can find our code in: \url{https://github.com/rohban-lab/Shaeiri_submitted_2020}
	
	\section*{Acknowledgement}
	We would like to thank Soroosh Baselizadeh, Hossein Yousefi Moghaddam, and Zeinab Golgooni for their insightful comments and reviews of this work.
	
	\break
	
	\nocite{langley00}
	
	\bibliography{example_paper}
	\bibliographystyle{icml2020}

	\clearpage
	\onecolumn
	\let\cleardoublepage\clearpage
	\appendix
	\appendixpage
	
	
	
	\section{Proofs} \label{Appendix_a}	
	
	\begin{definition_a}
		$P_{D, p, \epsilon}$ : The general problem of finding a $(p, \epsilon)$-optimal robust classifier on the distribution $D$, under the perturbation set $B_p^{\epsilon}(x)$.
		An algorithmic solution to this problem would take as input $D$, $p$ and $\epsilon$ and outputs a classifier $h : \mathbb{R}^d \rightarrow L$ that is $(p, \epsilon)$-optimal robust.
	\end{definition_a}

	\begin{definition_a}
		$Q_{G(V, E)}$: The problem of finding the maximum distance$-3$ independent set in an unweighted undirected graph $G(V, E)$. We call a subset of nodes $S \subseteq V$ a distance$-d$ independent set if $~\forall u_1, u_2 \in S, u_1 \neq u_2 : dist_G(u_1, u_2) \geq d$, where $dist_G(u_1, u_2)$ is length of the shortest path between $u_1$ and $u_2$ on the graph $G$. The goal of this problem is to find a maximum cardinality distance$-3$ independent set of graph $G(V, E)$.
	\end{definition_a}
	\begin{lemma_a} \label{red_lem}
		Every instance of $Q_{G(V, E)}$ can be reduced to an instance of $P_{D, p, \epsilon}$ in polynomial time.
	\end{lemma_a}
	\begin{proof}
		Let $|V| = k$ and consider an order $e_1, \ldots, e_{{k \choose 2}}$ of all possible edges on $G(V, E)$, where each possible edge is of the form $e_i = \{v_j, v_l\}$ and connects 2 nodes $v_j, v_l\ \in V, j \neq l$. We first construct a set of $k$ points in $d = {k \choose 2}$ dimensional space and then $D(x, y)$ be supported on these points. Let $M$ be $k \times d$ matrix, which is defined as:
		\begin{equation}
		M_{i, j} := 
		\begin{cases}
		0, & v_i \not\in e_j \\
		\lambda, &  e_j = \{v_i, v_l\} \text{ and } i > l \\
		-\lambda, & e_j = \{v_i, v_l\} \text{ and } i \leq l,
		\end{cases}
		\end{equation}
		and $\lambda$ is defined for any given $p, \epsilon \in \mathbb{R} , \ p, \epsilon > 0$ or $p = \infty$ as follows:
		\begin{equation}
		\lambda := 
		\begin{cases}
		\frac{\epsilon}{\sqrt[p]{2k - 4}}, & p \neq \infty \\
		\epsilon, & p = \infty.
		\end{cases}
		\end{equation}
		Note that the $\ell_p$ distance between every pair of rows in $M$ is greater than $\epsilon$. Now, based on $M$, we define a new matrix $M^{\prime}$, where if the edge $e_i \in E$ and $e_i = (v_j, v_l)$, we set $M^{\prime}_{j, i} = |M_{j, i}|$ and $M^{\prime}_{l, i} = |M_{l, i}|$. Note that in the matrix $M^{\prime}$, two rows whose corresponding vertices are not connected still remain greater than $\epsilon$ apart with respect to the $\ell_p$ norm. However, two rows that are connected in the graph $G(V, E)$ would now have an $\ell_p$ distance of exactly $\epsilon$.
				
		Next, let $L = \{ 1, 2, \ldots , k \}$ be the set of labels. We define discrete probability distributions $D : \mathbb{R}^d \times L \rightarrow \mathbb{R}$ on points defined as rows of $M^{\prime}$.
		\begin{equation}
		D(x = M^{\prime}_i, y = j) = 1/k \times \mathbb{I}(i = j)
		\end{equation}
		To find a $(p, \epsilon)$-optimal robust classifier on the mentioned $D$, which is supported on a finite set of points $X := \{M^{\prime}_1, \ldots, M^{\prime}_k \}$, we have to solve the following optimization problem:
		\begin{equation}
		\argmin_{h : \mathbb{R}^d \rightarrow L} R^{(adv)}_{p, \epsilon}(h, p)
		\end{equation}
		\begin{equation*}
		= \argmin_{h : \mathbb{R}^d \rightarrow L} \mathbb{E}_{x, y \ \sim D} \left\{ \mathbb{I}(\exists \delta \in B_{p}^{\epsilon} : h(x + \delta) \neq y) \right\}
		\end{equation*}
		\begin{equation*}
		= \argmin_{h : \mathbb{R}^d \rightarrow L} \sum_{x_i \in X, ~ 1 \leq y_j \leq k} \mathbb{I}\left(\exists \delta \in B_{p}^{\epsilon} : h(x + \delta) \neq y_j \right)  D(x_i, y_j)
		\end{equation*}
		\begin{equation*}
		= \argmax_{h : \mathbb{R}^d \rightarrow L} \sum_{x_i \in X, ~ 1 \leq y_j \leq k} \mathbb{I}\left(\forall \delta \in B_p^{\epsilon} : h(x_i + \delta) = y_j \right) D(x_i, y_j)
		\end{equation*}
		By our construction of $D$, this would be equivalent to:
		\begin{equation}
		\argmax_{h : \mathbb{R}^d \rightarrow L} \sum_{x_i \in X} \mathbb{I}\left(\forall \delta \in B_p^{\epsilon} :  h(x_i + \delta) = y_i \right)
		\end{equation}
		\begin{equation*}
		= \argmax_{h : \mathbb{R}^d \rightarrow L} \sum_{x_i \in X} \mathbb{I}\left(\forall \delta \in B_p^{\epsilon} :  h(x_i + \delta) = i \right)
		\end{equation*}
		Note that by our construction of $X$, there is an edge between $v_i$ and $v_j$ iff $\| x_j - x_i \|_p \leq \epsilon$. Hence, this optimization is equivalent to:
		\begin{equation}
		\argmax_{h : V \rightarrow L} \sum_{v_i \in V} \mathbb{I}\left(\forall u \in \text{neighbors}(v_i) : h(u) = i \text{ and } h(v_i) = i \right),
		\end{equation}
		
		where $\text{neighbors}(v)$ is the set of all nodes connected to the $v$ in the graph $G$. 		
		Note that for the $i$-th term in the sum to be one, (1) $v_i$ should be labeled as $i$ by $h$, and (2) all neighbors of $v_i$ in the graph $G(V, E)$ should be labeled as $i$. Therefore, for both the $i$-th and $j$-th term in the sum to be one, they should have a graph distance of at least 3. Also, if $v_i$ and $v_j$ have a graph distance of at least 3 then, they can both be one in the sum. As a result, the last optimization would be equivalent to:
		\begin{equation}
		\argmax_{U \subseteq V} \sum_{u \in U} \mathbb{I}(\forall v \in U, ~ v \neq u : dist_G(u, v) \geq 3).
		\end{equation}
		Therefore, the last optimization would equivalently find the set of vertices $U \subseteq V$ with maximum cardinality such that, all selected vertices have a distance of at least 3 in the graph. As a result, the problem of finding a $(p, \epsilon)$-optimal robust classifier on $D$ would yield a solution to the maximum distance$-3$ independent set for the input graph $G(V, E)$.
	\end{proof}
	
	\begin{theorem_a}
		The problem of finding the $(p, \epsilon)$-optimal robust classifier on the distribution $D$, $P_{D, p, \epsilon}$, is an $\mathcal{NP}$-hard problem.
		\qedhere
	\end{theorem_a}
	\begin{proof}
		As shown in Lemma \ref{red_lem}, the maximum distance$-3$ independent set problem can be reduced to $P_{D, p, \epsilon}$ in polynomial time. The former is a well-known $\mathcal{NP}$-hard problem, which implies that $P_{D, p, \epsilon}$ would be $\mathcal{NP}$-hard too.
		\qedhere
	\end{proof}

	\begin{remark_a}
		For the special case of $p = 0$ in Lemma \ref{red_lem}, one can consider $\epsilon = {k \choose 2} - 1 = d - 1$ and define the $k \times d$ matrix $M$ as: $M_{i, j} := (i - 1) \times d + j$. Note that the $\ell_0$ distance between every pair of rows in $M$ is exactly $d$. Now, based on $M$, we can again define a new matrix $M^{\prime}$, where if the edge $e_i \in E$ and $e_i = (v_j, v_l)$, we set $M^{\prime}_{j, i} = \min (M_{j, i}, M_{l, i})$ and $M^{\prime}_{l, i} = \min (M_{j, i}, M_{l, i})$. Note that in the matrix $M^{\prime}$, two rows whose corresponding vertices are not connected still remain exactly $d$ apart with respect to the $\ell_0$ norm. However, two rows that are connected in the graph $G(V, E)$ would now have an $\ell_0$ distance of exactly $d - 1$.
	\end{remark_a}	
	\begin{remark_a}
		The most general case of the perturbation set can be defined by using a relation $Re \subseteq \mathbb{R}^d \times \mathbb{R}^d$. Let $x \in \mathbb{R}^d$ then, the perturbation set is defined as: $B(x) := \left\{ x^{\prime} \in \mathbb{R}^d \; : \;
		(x, x^{\prime}) \in Re \right\}$. In this general case making a relation $Re$ using graph $G(V, E)$, and specifying coordinates of the points is trivial.
	\end{remark_a}
	
	\begin{theorem_a}
		Let $(x | y) \sim N(\mu_y, \sigma^2 I)$, and $y \in \{-1, 1\}$ with equal probabilities, then assuming that the $(2, \epsilon)$-optimal robust classifier has a continuous decision boundary, $f(x) = 2 \mathbb{I}(w^{\top} x \geq 0) - 1$ is $(2, \epsilon)$-robust , where $w = \mu_1 - \mu_{-1}$. Furthermore, the optimal Bayes and optimal robust classification error rates are $\Phi\left(\frac{-\| \mu_1 - \mu_{-1} \|_2}{2\sigma} \right)$, and $\Phi\left(\frac{-\| \mu_1 - \mu_{-1} \|_2 + 2\epsilon}{2\sigma} \right)$, respectively, where $\Phi$ is the cumulative distribution function for the standard normal distribution.
	\end{theorem_a}
	\begin{proof}
		A binary classifier can be expressed as {$h: \mathbb{R}^d \to \{-1, 1\}$.}
		We aim to find the optimal robust classifier $h^{\star (adv)}$:
		\begin{equation}
		h^{\star (adv)} = \argmin_{h} \mathbb{E}_{X, Y} \left\{ \max_{\|\delta\|_2 \leq \epsilon} |h(X + \delta) - Y|\right\},
		\end{equation}
		where $Y$ is the true label of the data point $X$.
		
		Let $f_1$ and $f_{-1}$ denote the conditional densities of $X$ given $Y$, $N(\mu_1, \sigma^2 I )$ and $N(\mu_{-1}, \sigma^2 I)$, accordingly. Then, we have:
		
		\begin{equation}
		\mathbb{P}(Y = i | X = x) = \frac{f_i (x)}{f_1(x) + f_{-1}(x)},\qquad i \in \{1, -1\},
		\end{equation}
		and we get:
		\begin{equation} \label{base}
		\begin{split} 
		& \mathbb{E}_{X, Y} \left\{ \max_{\|\delta\|_2 \leq \epsilon} |h(X + \delta) - Y|\right\} \\ & = \mathbb{E}_{X} \Biggl\{\frac{f_{1}(x)}{f_1(x) + f_{-1}(x)} \; \max_{\|\delta\|_2 \leq \epsilon}  |h(x + \delta) -1|\; + \;  \\
		&  ~~~~~~~~~~~~~~~~~~~~~~~ \frac{f_{-1}(x)}{f_1(x) + f_{-1}(x)} \; \max_{\|\delta\|_2 \leq \epsilon}  |h(x + \delta) + 1| \Biggr\} \\
		& = \frac{1}{2} \int_{\mathbb{R}^d} (f_1(x) \max_{\|\delta\|_2 \leq \epsilon} |h(x + \delta) - 1| \; + \;  f_{-1}(x)\max_{\|\delta\|_2 \leq \epsilon} |h(x + \delta) + 1|)dx
		\end{split}
		\end{equation}
		
		\begin{definition_a} \label{db}
			Decision Boundary of classifier $h$ is defined as
			\begin{equation}
			DB(h) := \left\{x ~ | ~ \exists C > 0 ~ \forall 0 < \epsilon \leq C, ~ \; \exists u \in \mathbb{R}^d,
			~~ u \neq 0, ~~ \|u\|_2 \leq \epsilon, \; h(x+u) \neq h(x)	
			\right\}
			\end{equation}
			In what follows, we focus only on the candidate classifiers that have a continuous decision boundary. 
		\end{definition_a}	
		
		\begin{definition_a} \label{cr}
			$\beta_{\epsilon}$(x) is defined as the set of points that are within $\ell_2$ ball of radius $\epsilon$ around $x$, excluding the outer boundary. Also, the critical region of $h$, denoted as $CR(h)$, is defined as the set of points that can be misclassified by adding a perturbation of length no more than $\epsilon$. 
			\begin{equation} \label{main_equ}
			CR(h) := \left\{x ~ | ~ \exists \delta \in \mathbb{R}^d, \; ~ \|\delta\|_2 \leq \epsilon, ~ \; h(x) \neq  h(x + \delta)\right\}
			\end{equation}
		\end{definition_a}
		
		\begin{lemma_a}
			$CR(h) = \bigcup\limits_{x \in DB(h)} \beta_{\epsilon} (x)$ 
		\end{lemma_a}
		
		\begin{proof}
			Suppose that $x \in CR(h)$. Therefore, there exists $\delta \text{ such that } \|\delta\|_2 \leq \epsilon$ and $h(x) \neq h(x + \delta)$. Since the label of the data points, $x$ and $x + \delta$ are different and as we assumed that $DB(h)$ is continuous, there must exist  $k \in (0, 1)$ such that $x + k\delta \in DB(h)$.
			Note that 
			\begin{center}
				$ \|x + k\delta - x \|_2 = \|k\delta\|_2 = k \|\delta\|_2 < \|\delta\|_2 \leq \epsilon \implies x + k\delta \in \beta_{\epsilon}(x)$, \\
			\end{center}
			and according to Definition \ref{cr}, 
			$CR(h) \subseteq \bigcup\limits_{x \in DB(h)} \beta_{\epsilon} (x)$. \\ 
			
			Now suppose that $x \in \bigcup\limits_{x \in DB(h)} \beta_{\epsilon} (x)$. Then, there exists $x_0 \in DB(h)$ such that $x \in \beta_{\epsilon}(x_0)$. 
			But note that
			$$x \in \beta_{\epsilon}(x_0)  \implies \|x-x_0\|_2 < \epsilon$$
			$$x_0 \in DB(h) \implies \exists u: \|u\|_2 \leq \epsilon - \|x - x_0\|_2, h(x_0 + u) \neq h(x_0)$$
			
			As a result, $\|x - (x_0 + u) \|_2 \leq  \|x - x_0\|_2 + \|u\|_2 \leq \epsilon $. 
			
			Note that as $\|x-x_0\|_2 < \epsilon$, $h(x) \neq h(x + (x_0 - x)) = h(x_0)$, if $h(x_0) \neq h(x)$, $x \in CR(h)$.
			Otherwise,
			$$h(x_0) = h(x), \quad h(x_0 + u) \neq h(x_0)  \implies h(x_0 + u) \neq h(x),$$
			$$ \|x - (x_0 + u)\|_2 \leq \epsilon \implies x \in CR(h).$$
			
			So, in both cases, $\bigcup\limits_{x \in DB(h)} \beta_{\epsilon} (x) \subseteq CR(h)$.	
		\end{proof}
		
		According to the Definition \ref{cr}, for all $x \in CR(h)$, there exist two points like $x_0 = x$, $x_1$ within $\epsilon$ distance of $x$, such that $h(x_0) = 1$, $h(x_1) = -1$. Therefore,
		$\max_{\|\delta\|_2 \leq \epsilon} |h(x + \delta) - 1| = \max_{\|\delta\|_2 \leq \epsilon} |h(x+\delta) + 1 | = 2$. As a result, we have
		\begin{multline} \label{cr_equ}
		\frac{1}{2} \int_{CR(h)} (f_1(x) \max_{\|\delta\|_2 \leq \epsilon} |h(x + \delta) - 1|   +  f_{-1}(x)\max_{\|\delta\|_2 \leq \epsilon} |h(x + \delta) + 1|) dx = \int_{CR(h)} (f_1(x) + f_{-1}(x)) dx.
		\end{multline}
		
		\begin{definition_a} \label{sr}
			Safe Region of the class $i$, denoted as $SR_i(h)$, is defined as the set of points that are labeled as $i$ by the classifier $h$, and cannot be misclassified when a perturbation of size less than or equal to $\epsilon$ is added.
		\end{definition_a}
		According to the definition above, for all $x_0 \in SR_i(h)$, we get 
		\begin{equation*} 
		\forall \delta ~ \|\delta\|_2 \leq \epsilon \quad h(x_0) = h(x_0 + \delta) = i.
		\end{equation*}
		As a result, for such $x_0$, $\max_{\|\delta\|_2 \leq \epsilon} |h(x_0 + \delta) - 1| = |i - 1|$ and $\max_{\|\delta\|_2 \leq \epsilon} |h(x_0 + \delta) + 1| = |i + 1|$. Therefore, the value of  
		\begin{equation*}
		\frac{1}{2} \int_{SR_i(h)} f_1(x) \max_{\|\delta\|_2 \leq \epsilon} |h(x + \delta) - 1| \; + \; f_{-1}(x)\max_{\|\delta\|_2 \leq \epsilon} |h(x + \delta) + 1|\; dx
		\end{equation*}
		for $i = 1, -1$ is 
		\begin{equation} \label{sr_equ}
		\int_{SR_1(h)} f_{-1}(x) dx, ~~~~ \text{ and } ~~~~ \int_{SR_{-1}(h)} f_{1}(x) dx,
		\end{equation}
		respectively.
		
		As $\mathbb{R}^d = CR(h) \bigcup SR_1(h) \bigcup SR_{-1}(h)$, and according to Eqs. \ref{base}, \ref{cr_equ}, and \ref{sr_equ}, we have
		\begin{multline}
		\mathbb{E}_{X, Y} \left\{ \max_{\|\delta\|_2 \leq \epsilon} |h(x + \delta) - y|\right\} = \int_{CR(h)}(f_1(x) + f_{-1}(x)) dx \;+ \; \int_{SR_1(h)}f_{-1}(x)dx \\
		\; + \;  \int_{SR_{-1}(h)}f_{1}(x)dx =  2 - \int_{SR_1(h)} f_1(x) dx - \int_{SR_{-1}(h)} f_{-1}(x) dx 
		\end{multline}
		
		So the solution to optimal $\epsilon$ robust classifier in Eq. \ref{main_equ}, is equivalent to the solution of
		\begin{equation}
			\argmax_{h} \int_{SR_1(h)} f_1(x)dx + \int_{SR_{-1}(h)} f_{-1}(x)dx
		\end{equation}
		Let $E$ be a unit vector parallel to the line passing means of the normal distributions. That is, $E := \frac{\mu_1 - \mu_{-1}}{\|\mu_1 - \mu_{-1}\|_2}$. Define $SR_i'(h) = \{ x - i\epsilon E ~|~ x \in SR_i(h)\}$ and $f_i'$ as the density of the normal distributions with mean of $\mu_1 - i\epsilon E$ and covariance $\sigma^2 I$, for $i = 1, -1$. Then, we have 
		\begin{equation}
		\begin{split}
		\int_{SR_1(h)} & f_1(x) dx + \int_{SR_{-1}(h)} f_{-1}(x) dx \\ & = \int_{SR_1'(h)} f_1(x + \epsilon E) dx + \int_{SR_{-1}'(h)} f_{-1}(x - \epsilon E) dx  \\ 
		& =  \int_{SR_1'(h)} f_1'(x) dx + \int_{SR_{-1}'(h)} f_{-1}'(x) dx
		\end{split}
		\end{equation}
		
		\begin{lemma_a}
			$SR_1'(h) \bigcap SR_{-1}' (h) = \emptyset$.
		\end{lemma_a}
		\begin{proof}
			Otherwise, there exists $x \in  SR_1'(h), x \in SR_{-1}'(h)$, such that
			\begin{align*}
			x - \epsilon E \in SR_{-1}(h)  \quad \overset{\|\epsilon E\|_2 = \epsilon}{\implies} \quad h(x- \epsilon E) = h(x) = -1 \\			x + \epsilon E \in SR_{1}(h) \quad \overset{\|\epsilon E\|_2 = \epsilon}{\implies} \quad h(x + \epsilon E) = h(x) = 1,
			\end{align*} 		
			which is a contradiction.
		\end{proof}
		
		So every $ x \in \mathbb{R}^d$, belongs to at most one of $SR_1'(h)$ and $SR_{-1}'(h)$ and so, $SR_{-1}'(h) \subseteq \mathbb{R}^d \backslash SR_1'(h)$. This implies that
		\begin{equation}
		\int_{SR_1'(h)} f_1'(x) dx + \int_{SR_{-1}(h)} f_{-1}'(x) dx  \leq \int_{SR_1'(h)} f_1'(x) dx + \int_{ \mathbb{R}^d \backslash SR_1'(h)} f_{-1}'(x) dx 
		\end{equation}
		Now, let $DB^\star$ be the hyperplane perpendicular to $E$, $h^\star$ be the classifier whose decision boundary is $DB^\star$, and $h^\star(\mu_1) = 1, h^\star(\mu_{-1}) = -1$. Then
		\begin{equation*}
		\forall x \text{ s.t. } h^\star(x) = i, \quad \|x - \mu_i \|_2 \leq \|x - \mu_{-i} \|_2 \\
		\end{equation*}

		As $f_1'$ and $f_{-1}'$ are densities of the normal distributions with the same covariance, for $\epsilon$ less than half of the distance between $\mu_1$ and $\mu_{-1}$, if $\|x - \mu_1\|_2 \leq \|x - \mu_{-1}\|_2$, then $f_1'(x) \geq f_{-1}'(x)$. Further, if $\|x - \mu_1\|_2 \geq \|x - \mu_{-1}\|_2$, then $f_1'(x) \leq f_{-1}'(x)$. \\
		
		Now define $C_i = \{x | h^\star(x)  = i\}$. By definition, $C_1 \bigcup C_{-1} = \mathbb{R}^d$. Then, $\forall x \in C_i, \quad f_i'(x) \geq f_{-i}'(x)$. For any classifier, we have
		\begin{equation}
		\begin{split}
		\int_{SR_1'(h)} & f_1'(x) dx + \int_{ \mathbb{R}^d \backslash SR_1'(h)} f_{-1}'(x) dx  \\ & =  \int_{SR_1'(h) \bigcap C_1} f_1'(x)dx +  \int_{SR_1'(h) \bigcap C_{-1}} f_1'(x)dx \\
		&  +  \int_{\mathbb{R}^d \backslash SR_{1}'(h) \bigcap C_1} f_{-1}'(x)dx +  \int_{\mathbb{R}^d \backslash  SR_{1}'(h) \bigcap C_{-1}} f_{-1}'(x)dx \\ & \leq  \int_{SR_1'(h) \bigcap C_1} f_1'(x)dx + 
		\int_{SR_1'(h) \bigcap C_{-1}} f_{-1}'(x)dx \\ &  +  \int_{\mathbb{R}^d \backslash SR_{1}'(h) \bigcap C_1} f_{1}'(x)dx +  \int_{\mathbb{R}^d \backslash SR_{1}'(h) \bigcap C_{-1}} f_{-1}'(x)dx  \\
		& = \int_{C_1} f_1'(x) dx + \int_{C_{-1}} f_{-1}'(x) dx
		\end{split}
		\end{equation}
		
		Finally, note that for the classifier $h^\star$, the RHS and LHS of the above inequality are equal. Therefore, $h^\star$ is an optimal $\epsilon$ robust classifier and the optimal robust classification error would clearly be $\Phi\left(\frac{-\| \mu_1 - \mu_{-1} \|_2 + 2\epsilon}{2\sigma} \right)$.\\
	\end{proof}
	
	\begin{remark_a}
		\cite{ilyas2019adversarial} has studied the problem of the optimal robust classifier on a binary classification task where data is driven from Gaussian distributions. However, they have a strong assumption on the hypothesis space. Specifically, they focused on linear classifiers. Here, we do not assume anything about the hypothesis space other than continuity of the decision boundary.
	\end{remark_a}
		
	\begin{fact_a} \label{ffact}
		Assume that $\theta^{(1)}$ and $\theta^{(-1)}$ are different in $k \leq d$ dimensions. We call $k$ the ``effective dimensionality" and we get $k = \| \theta^{(1)} - \theta^{(-1)}\|_0$. We can restrict all $x$, $\theta^{(1)}$, and $\theta^{(-1)}$ only to the $k$ effective dimensions. We call $x^{\prime}$, $\theta^{\prime (1)}$, and $\theta^{\prime (-1)}$ the restriction of these vectors to the ``effective dimensions". Similarly, we call $x^{\prime\prime}$, $\theta^{\prime \prime(1)}$, and $\theta^{\prime \prime(-1)}$ the restriction of these vectors to ``non-effective dimensions".
		$ $\newline
		Note that $\theta^{\prime(1)} = - \theta^{\prime(-1)}$ and $\theta^{\prime\prime(1)} = \theta^{\prime\prime(-1)}$, and
		\begin{equation} \label{sum_k}
		\| x^{\prime} - \theta^{\prime(1)} \|_0 + \| x^{\prime} - \theta^{\prime(-1)} \|_0 = k.
		\end{equation}
	\end{fact_a}
	
	\begin{theorem_a} \label{lemma_bayes}
		Let's assume that $(x, y) \sim \text{Bernoulli} \ (\theta^{(1)}, \theta^{(-1)}, t)$. Then, 1. the optimal Bayes' classifier $h^\star : \{-1, 1\}^d \rightarrow \{-1, 1\}$ on this distribution is of the form:	
		\begin{equation}
		h^\star(x) = 
		\begin{cases}
		1, & \| x - \theta^{(1)} \|_0 \leq \| x - \theta^{(-1)} \|_0 \\
		-1, & \text{otherwise},
		\end{cases}
		\end{equation}
		and 2. if $k$ is even, the optimal classification error rate is:	
		\begin{equation}
		\sum_{i = 0}^{k/2} {k \choose i} (1/2 + t)^{i} (1/2 - t)^{k - i} - 1/2 {k \choose k/2} (1/2 + t)^{k/2} (1/2 - t)^{k/2}
		\end{equation}
		
		\break
		
		Otherwise, if $k$ is odd, the classification optimal error rate would be:
		\begin{equation}
		\sum_{i = 0}^{\lfloor k/2 \rfloor} {k \choose i} (1/2 + t)^{i} (1/2 - t)^{k - i},
		\end{equation}	
		where $k := \| \theta^{(1)} - \theta^{(-1)} \|_0$.
	\end{theorem_a}
	\begin{proof}
		1. We start with the equation of the Baye's optimal classifier as an optimal classifier on the given distribution: 
		\begin{equation}
		h^\star(x) = \argmax_{y \in \{ -1, 1 \}} ~ D(x, y)
		\end{equation}
		\begin{equation*}
		= \argmax_{y \in \{ -1, 1 \}} ~ 1/2 \times (1/2 + t)^{k - \| x - \theta^{(y)} \|_0} (1/2 - t)^{\| x - \theta^{(y)} \|_0}
		\end{equation*}
		\begin{equation*}
		= \argmax_{y \in \{ -1, 1 \}} ~ (1/2 + t)^{k - \| x - \theta^{(y)} \|_0} (1/2 - t)^{\| x - \theta^{(y)} \|_0}
		\end{equation*}
		\begin{equation*}
		= \argmax_{y \in \{ -1, 1 \}} ~ (1/2 + t)^{k - \| x^\prime - \theta^{\prime(y)} \|_0} (1/2 - t)^{\| x^\prime - \theta^{\prime(y)} \|_0}
		\end{equation*}
		That is:
		\begin{equation}
		h^\star(x) = 
		\begin{cases}
		1, & (1/2 + t)^{k - \| x^\prime - \theta^{\prime(1)} \|_0} (1/2 - t)^{\| x^\prime - \theta^{\prime(1)} \|_0} \geq \\  & ~~~~~~~~~~ (1/2 + t)^{k - \| x^\prime - \theta^{\prime(-1)} \|_0} (1/2 - t)^{\| x^\prime - \theta^{\prime(-1)} \|_0} \\
		-1, & \text{otherwise}
		\end{cases}
		\end{equation}
		
		By using the Eq. \ref{sum_k} in the above classification rule:
		
		\begin{equation}
		h^\star(x) = 
		\begin{cases}
		1, & (1/2 + t)^{k - \| x^\prime - \theta^{\prime(1)} \|_0} (1/2 - t)^{\| x^\prime - \theta^{\prime(1)} \|_0} \geq \\  & ~~~~~~~~~~ (1/2 + t)^{\| x^\prime - \theta^{\prime(1)} \|_0} (1/2 - t)^{k - \| x^\prime - \theta^{\prime(1)} \|_0} \\
		-1, & \text{otherwise}
		\end{cases}
		\end{equation}
		
		By rearranging this rule, we finally arrive at:
		\begin{equation}
		h^\star(x) = 
		\begin{cases}
		1, & {\| x^\prime - \theta^{\prime(1)} \|_0} \leq k/2 \\
		-1, & \text{otherwise,}
		\end{cases}
		\end{equation}	
		which by applying the Eq. \ref{sum_k} again is the same as:
		\begin{equation}
		h^\star(x) = 
		\begin{cases}
		1, & {\| x^\prime - \theta^{\prime(1)} \|_0} \leq {\| x^\prime - \theta^{\prime(-1)} \|_0} \\
		-1, & \text{otherwise}
		\end{cases}
		\end{equation}
		And finally:
		\begin{equation}
		h^\star(x) = 
		\begin{cases}
		1, & {\| x - \theta^{(1)} \|_0} \leq {\| x - \theta^{(-1)} \|_0} \\
		-1, & \text{otherwise}
		\end{cases}
		\end{equation}
		2. The classification error rate for $h^\star$ can now be calculated as follows:
		\begin{equation}
		R(h^\star, D) = \mathbb{E}_{x, y \ \sim D} \left\{ \mathbb{I}(h^\star(x) \neq y) \right\}
		\end{equation}
		\begin{equation*}
		= \sum_{x, y} \mathbb{I}(h^\star(x) \neq y) \times D(x, y)
		\end{equation*}
		\begin{equation*}
		= \sum_{x : h^\star(x) = 1} D(x, -1) \ + \sum_{x : h^\star(x) = -1} D(x, 1)
		\end{equation*}
		By plugging in $h^\star$ and $D(x, y)$, we would get:
		\begin{multline}
		~~~~~~~~~~~~~~~~~~~~~~~~~~~~ = 1/2 \sum_{x : \| x^\prime - \theta^{\prime(1)} \|_0  \leq k/2} (1/2 + t)^{\| x^\prime - \theta^{\prime (1)} \|_0} (1/2 - t)^{k - \| x^\prime - \theta^{\prime (1)} \|_0} \\ 
		+ 1/2 \sum_{x : \| x^\prime - \theta^{\prime(1)} \|_0 > k/2} (1/2 + t)^{k - \| x^\prime - \theta^{\prime (1)} \|_0} (1/2 - t)^{\| x^\prime - \theta^{\prime (1)} \|_0} 
		\end{multline}
		
		If $k$ is even, this would simplify to:
		\begin{equation}
		1/2 \sum_{i = 0}^{k/2} {k \choose i} (1/2 + t)^{i} (1/2 - t)^{k - i} + \ 1/2 \sum_{i = k/2 + 1}^{k} {k \choose i} (1/2 + t)^{k - i} (1/2 - t)^{i}
		\end{equation}
		
		\begin{equation}
		= \sum_{i = 0}^{k/2} {k \choose i} (1/2 + t)^{i} (1/2 - t)^{k - i} - 1/2 {k \choose k/2} (1/2 + t)^{k/2} (1/2 - t)^{k/2}
		\end{equation}
		Otherwise, if $k$ is odd, the error rate would be:
		\begin{equation}
		\sum_{i = 0}^{\lfloor k/2 \rfloor} {k \choose i} (1/2 + t)^{i} (1/2 - t)^{k - i}
		\end{equation}
	\end{proof}
	
	\noindent
	\begin{remark_a}
		Note that the final equations of the optimal classification error rate for this distribution can be simply rewritten by using the Hypergeometric functions.
	\end{remark_a}
	
	
	\begin{theorem_a}
		Let $(x, y) \sim \text{Bernoulli}(\theta^{(1)}, \theta^{(-1)}, t)$.
		Let $B_0^{\epsilon}(x)$ be the perturbation set.
		Then, under the assumption \ref{symm_asm}, 1. the $(0, \epsilon)$-optimal robust classifier $h^\star : \{-1, 1\}^d \rightarrow \{-1, 1\}$ on this distribution is of the form:
		\begin{equation}
		h^{\star (adv)}(x) = 
		\begin{cases}
		1, & \| x^\prime - \theta^{\prime (1)} \|_0 \leq s \\
		-1, & \text{otherwise},
		\end{cases}
		\end{equation}
		where $s \in \{0, 1, ..., k\}$ is a threshold, and $x^\prime$, and $\theta^{\prime (1)}$ are the restrictions of $x$, and $\theta^{(1)}$ to the effective dimensions, defined in theorem \ref{lemma_bayes_main}, respectively,
		and 2. the classification $(0, \epsilon)$-optimal error rate would be:
		\begin{equation}
		R_{0, \epsilon}^{(adv)}(h, D) =
		1/2 \sum_{i = 0}^{s + \epsilon} {k \choose i} (1/2 + t)^{i} (1/2 - t)^{k - i} + 1/2 \sum_{i = 0}^{k - s + \epsilon - 1} {k \choose i} (1/2 + t)^{i} (1/2 - t)^{k - i}
		\end{equation}
	\end{theorem_a}
		
	\begin{proof}
		
		
		
		
		1. Let $k$ be an effective dimension according to Fact \ref{ffact}. Let $h$ be a $(0, \epsilon)$-optimal robust classifier. Let $S_1, S_{-1} \subset \{-1, 1\}^d$ denote all the points that are assigned to the classes $1$ and $-1$ by $h$, and have $\ell_0$ distance of no more than $\ceil*{\frac{k}{2}}$ to $\theta^{\prime (1)}$ and $\theta^{\prime (-1)}$, respectively. Further, let $x^\star_1 := \arg\max_{x \in S_1} \| x - \theta^{\prime (1)} \|_0$, and \hbox{$x^\star_{-1} := \arg\max_{x \in S_{-1}} \| x - \theta^{\prime (-1)} \|_0$}. And also, let $d_{1} = \| x^\star_{1} - \theta^{\prime (1)} \|_0$ and, $d_{-1} = \| x^\star_{-1} - \theta^{\prime (-1)} \|_0$.  We claim that for such an optimal robust classifier $h$, all points $x \in \{-1, 1\}^d$, which $\| x - \theta^{\prime (1)} \|_0 \leq d_{1}$ can be labelled as $1$, and all points $x \in \{-1, 1\}^d$, which $\| x - \theta^{\prime (-1)} \|_0 \leq d_{-1}$ can be labelled as $-1$. To show this, consider a set of points $\bar{S_{1}} = \{x \in \{-1, 1\}^d \ | \ h(x) = -1, \| x - \theta^{\prime (1)} \|_0 \leq d_{1} \}$. Then, by changing their labels to $+1$: 1. We will not destroy the robustness of any points in $S_{1}$, because they are all labelled as $+1$ by the $h$. 2. We will not destroy the robustness of the other points in $\{-1, 1\}^d - S_{1}$, because in fact $x^\star_{1}$ and all points $T_{1} = \{ x \in \{-1, 1\}^d : \| x - \theta^{\prime (1)} \|_0 = d_{1} \}$ have the same label $+1$ using the assumption \ref{symm_asm}. Also, $\ell_0$ distance between $\bar{S_{1}}$ and $\{-1, 1\}^d - S_{1}$ are greater or equal than the $\ell_0$ distance between $T_{1}$ and $\{-1, 1\}^d - S_{1}$. 3. Because for $\bar{S_{1}}$ the probability of being $+1$ is greater than being $-1$, this would yield a classifier that does not have a worse adversarial error rate compared to $h$. So, overall we can change these points label and similarly the points in $\bar{S}_{-1}$ without loss of accuracy.

		 Therefore, the $(0, \epsilon)$-optimal robust classifier basically thresholds $\| x^{\prime} - \theta^{\prime (1)} \|_0$. Notably, the thresholds can be either $0$ or $k$, which means that all points are labeled to either $+1$ or $-1$. The minimum adversarial error for $p = 0$ and $\epsilon$ is given by:
		\begin{equation}
		\min_{0 \leq s \leq k} 1/2 \sum_{i = 0}^{s + \epsilon} {k \choose i} (1/2 + t)^{i} (1/2 - t)^{k - i} + 1/2 \sum_{i = 0}^{k - s + \epsilon -1} {k \choose i} (1/2 + t)^{i} (1/2 - t)^{k - i}
		\end{equation}
		For specifying the exact threshold, one can evaluate the adversarial error rate using a different value of $k$ (Supplementary Figure in Sec. \ref{FIG}).
		%
	\end{proof}

	\begin{lemma_a}
		Let $B_p^{\epsilon}(x)$ be a perturbation set, and $h : \mathbb{R}^d \rightarrow L$ be a classifier.
		Then, $R^{(adv)}_{p, \epsilon}(h, D) \geq R(h, D)$. 		
	\end{lemma_a}
	\begin{proof}
		
		\begin{equation}
		R_{p, \epsilon}^{(adv)}(h, D) = \mathbb{E}_{x, y \ \sim \ D} \left\{ \mathbb{I}(\exists \delta \in B_{p}^{\epsilon} : h(x + \delta) \neq y) \right\}
		\end{equation}
		
		It is obvious that $B_p^{0}(x) \subseteq B_p^{\epsilon}(x)$ then:
		
		\begin{equation}
		\mathbb{E}_{x, y \ \sim \ D} \left\{ \mathbb{I}(\exists \delta \in B_{p}^{\epsilon} : h(x + \delta) \neq y) \right\} \geq \mathbb{E}_{x, y \ \sim \ D} \left\{ \mathbb{I}(\exists \delta \in B_{p}^{0} : h(x + \delta) \neq y) \right\}
		\end{equation}
		
		\begin{equation}
		= \mathbb{E}_{x, y \ \sim \ D} \left\{ \mathbb{I}(h(x) \neq y) \right\} = R(h, D)
		\end{equation}
		
	\end{proof}

	\begin{theorem_a}
		Let $\text{Bernoulli} \ (\theta^{(1)}, \theta^{(-1)}, t)$ be a Bernoulli model.
		Let $(x, y) \sim \text{Bernoulli}(\theta^{(1)}, \theta^{(-1)}, t)$.
		Let $B_\infty^{\epsilon}(x)$ be a perturbation set.
		Then, 1. the $(\infty, \epsilon)$-optimal robust classifier $h^{\star (adv)} : \{-1, 1\}^d \rightarrow \{-1, 1\}$ on this distribution for $\epsilon < 1$ is of the form: 
		\begin{equation}
		h^{\star (adv)}(x) = 
		\begin{cases}
		1, & \| T(x) - \theta^{(1)} \|_0 \leq \| T(x) - \theta^{(-1)} \|_0 \\
		-1, & \text{otherwise},
		\end{cases}
		\end{equation}
		where $T(x) = (\sign(x_1), \ldots, \sign(x_d))$ and $\sign(r)$ is $1$ if $r \geq 0$ and is $-1$ otherwise,
		and 2. the classification $(\infty, \epsilon)$-optimal error rate would be:
		\begin{equation}
		R_{\infty, \epsilon}^{(adv)}(h, D) = 
		\begin{cases}
		R(h^{\star}, D), & \epsilon < 1 \\
		1/2, & \text{otherwise},
		\end{cases}
		\end{equation}
		where $h^{\star}$ is the optimal Baye's classification rule.
	\end{theorem_a}

	\begin{proof}
		Note that if $\epsilon < 1$, no $\ell_\infty$ attack could change the sign of any dimension of $x$. Therefore, applying the function $T$ would recover the original clean input $x$. Therefore, the adversarial error rate of $h^\star \circ T$ would be the same as the error rate of $h^\star$. But adversarial error rate of any classifier is lower bounded by optimal Bayes error rate. Here, the function $h^\star \circ T$ achieves this lower bound. Hence, this classifier should be $(\infty, \epsilon)$-optimally robust. If $\epsilon \geq 1$, we can simply map all points of the distribution to $\overrightarrow{0}$. Therefore even the $(\infty, \epsilon)$-optimal robust classifier can not have better than random accuracy, which is $1/2$ here.
	\end{proof}
	
	\break
	
	\section{Different Settings for the Adversarial Training} \label{appendix_b}
	For the MNIST dataset, we tested a variety of settings to adversarially train a model on $\epsilon = 0.4$. However, it seems that training a robust network on $0.4$ is not feasible when weights are randomly initialized. It is notable that this fact was also observed by \cite{Chen_2017}.
	\begin{enumerate}
		\item We test \cite{pmlrv9glorot10a} initialization and all other settings are the same as the main text. 
		\item We test different settings of optimizers,  specifically, SGD with the learning rate from $\{0.1, 0.01, 0.001\}$ with and without momentum and Adam \cite{kingma2014adam} with the learning rate from $\{0.1, 0.01, 0.001\}$ and all other settings are the same as the main text.
		\item Also, we try to change the batch size from $\{16, 32, 64, 128, 256, 512, 10^4\}$ and all other settings are the same as the main text.
		\item We test standard ResNet-18 architecture \cite{he2015deep} with different settings of optimizers as mentioned below. We believe that this model can cover needed model capacity as discussed in \cite{Madry2017TowardsDL}.
		\item We test different settings of PGD \cite{Madry2017TowardsDL} by using different learning rates and step sizes and all other settings are the same as the main text. It is notable that by using FGSM, we can successfully train a model on $\epsilon = 0.4$. However, the model indeed suffers from the gradient obfuscation \cite{DBLP:journals/corr/abs-1802-00420}.
	\end{enumerate}

	\section{Evaluation Details} \label{appendix_c}
	When we use the training PGD as mentioned in the main text on the epsilon $= 0.4$ model, the attack leads to about $89\%$ adversarial accuracy.  Also, when we use the training PGD as mentioned in the main text on the epsilon = $0.5$ model, the attack leads to about $80\%$ adversarial accuracy!
	\begin{enumerate}
		\item FGSM \cite{Goodfellow2014ExplainingAH} is not a stronger attack than PGD \cite{Madry2017TowardsDL}. However, verifying that PGD is always stronger than FGSM on our models is essential. We verified that adversarial accuracy is always lower using training PGD on both models.
		\item Unbounded attacks should always lead to zero accuracies. We verify that by using the training PGD but in the unbounded setting, we can make an attack, which leads to zero accuracies on both models.
		\item PGD with random offset should not affect the model adversarial accuracy significantly. As the random offset is included in training PGD, attacking the model using random offset does not decrease the adversarial accuracy on models.
		\item Using many random restarts should not effect model adversarial accuracy significantly. We used training PGD with 10 random restarts. However, we did not observe any significant changes in the adversarial accuracy of both models.
		\item Using PGD with more steps should not affect adversarial accuracy significantly. We used the training PGD with 2000 steps as an attack but it did not affect models' adversarial accuracy significantly.
		\item By increasing the magnitude of the attack, adversarial accuracy should monotonically decrease. We verify that by using the training PGD and increasing the allowed magnitude of perturbations, the adversarial accuracy of both models monotonically decreases.
		\item The targeted PGD attack should not significantly change the adversarial accuracy. We verify that by using the PGD targeted attack with the same setting as in the training PGD, and considering the highest loss between 10 possible classes targets for each image \cite{gowal2019alternative}. Still, no significant changes are observed using this new attack on both models.
		\item A brute force attack should be an effective attack on the model. We use a batch of size 128  and $10^8$ samples of perturbation for each image. Specifically, we use Bernoulli distribution for sampling the perturbations. Each pixel of an image can change either $-\epsilon$ or $+\epsilon$. We verify that this method also is not effective in attacking both models.
		\item We also verify that the minimum needed size of distortion for decreasing the adversarial accuracy of models to zero, is higher in both models in comparison to the base model, assuming the training PGD as an attack.
		\item We also try to decrease our models' adversarial accuracy by using the black-box attack ZOO \cite{Chen_2017}. Specifically, we used the default parameters in the art library \cite{art2018}. But this attack is not also effective on both models. Here, we note that indeed being sure about the effect of this attack needs more hyper parameters tuning.
		\item Black box attacks should not outperform white-box attacks. We also verify our models' adversarial accuracy on black-box attacks is not lower than the reported adversarial accuracy. Specifically, for the black box attack, we used the training PGD on the model robust to the adversarial perturbations of size $0.3$.
		\item We also propose a new attack which we call ``semi-black box". In this attack, we used the base model for training on perturbations of sizes 0.4 and 0.5, as a model for generation of the adversarial examples. We used the same setting as in the black-box attacks. However, again this new attack is also not effective.
		\\
    	\\
    	\\
    	\\
    	\\
    	\\
    	\\
    	\\
    	\\
    	\\
    	\\
    	\\
    	\\
    	\\
    	\\
    	\\
    	\\
    	\\
    	\\
    	\\
    	\\
    	\\
    	\\
    	\\
    	\\
    	\\
    	\\
    	\\
    	\\
    	\\
    	\\
    	\\
    	\\
    	\\
    	\\
    	\\
    	\\
    	\\
    	\\
    	\\
    	\\
    	\\
    	\\
		
	\end{enumerate}

	\section{Supplementary Figures for the Proposed Method} \label{appendix_d}

	\begin{figure}[H]
		\centering
		\includegraphics[width=0.9\linewidth]{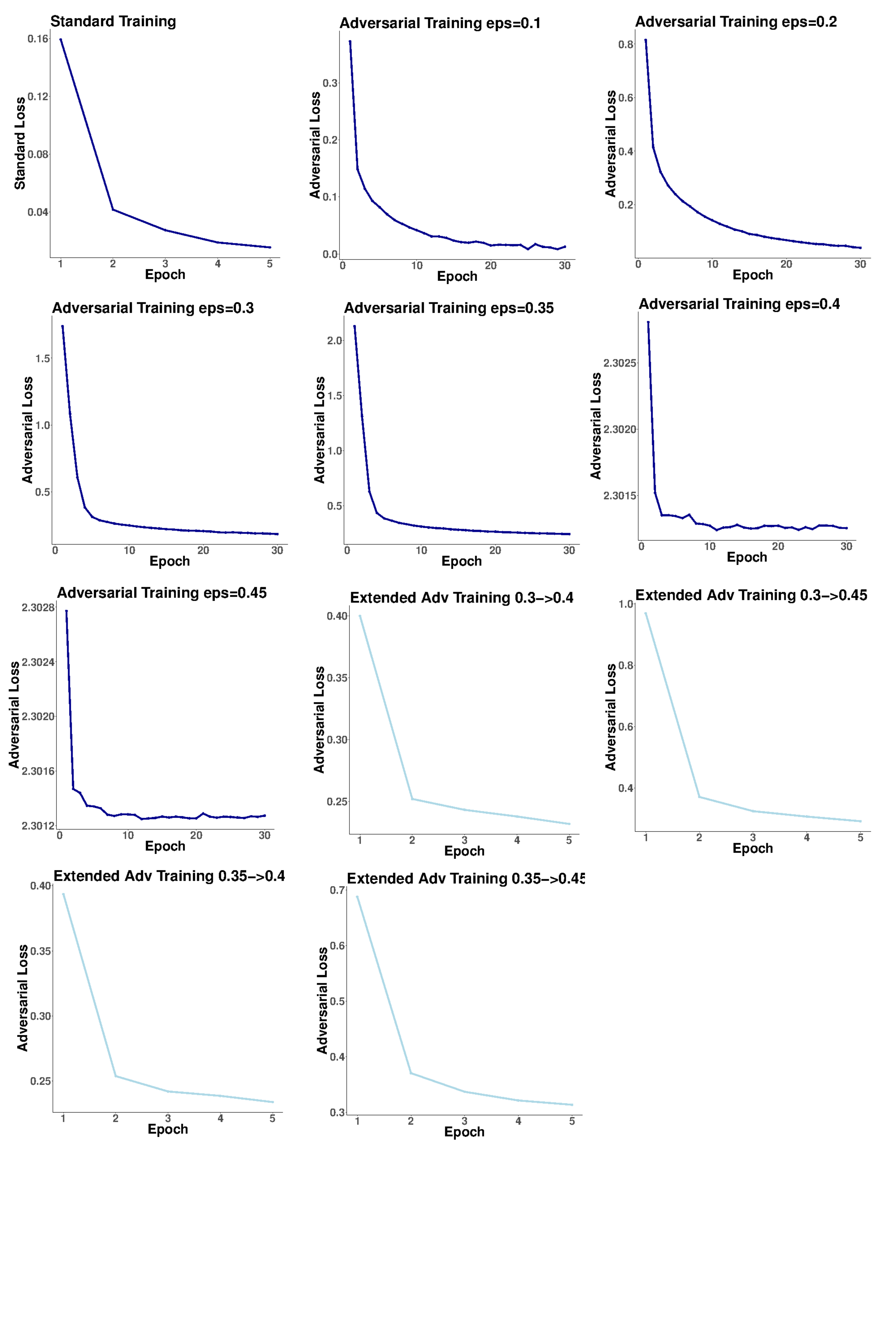}
		\vspace{-100pt}
		\caption{Training adversarial loss against the epoch number for various epsilons in the MNIST dataset. Adversarial training fails to decrease the training adversarial loss on epsilon greater than $\approx 0.35$. However, training such models is feasible using extended adversarial training.}
	\end{figure}
	
	\begin{figure}[H] 
		\centering
		\includegraphics[width=0.9\linewidth]{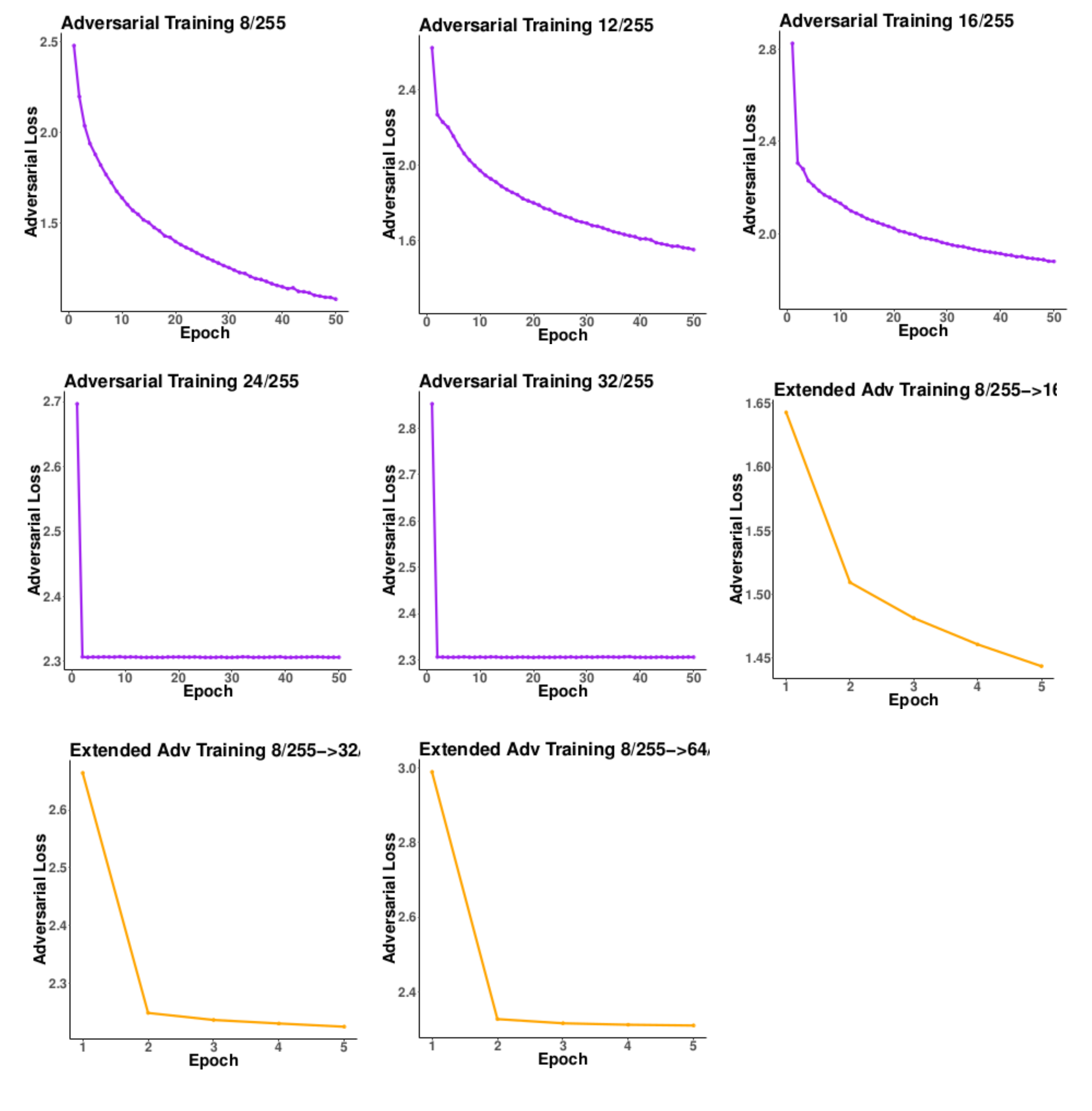}
		\caption{Training adversarial loss against the epoch number for various epsilons in the CIFAR10 dataset. Adversarial training fails to decrease the training adversarial loss on epsilon greater than $\approx \frac{16}{255}$. However, training such models is feasible using extended adversarial training.}
		\label{app_fig_cifar}
	\end{figure}


	\section{Omitted Figures for the Convergence of Adversarial Training} \label{appendix_e}
	
	\begin{figure}[H]
		\centering
		\includegraphics[width=0.8\linewidth]{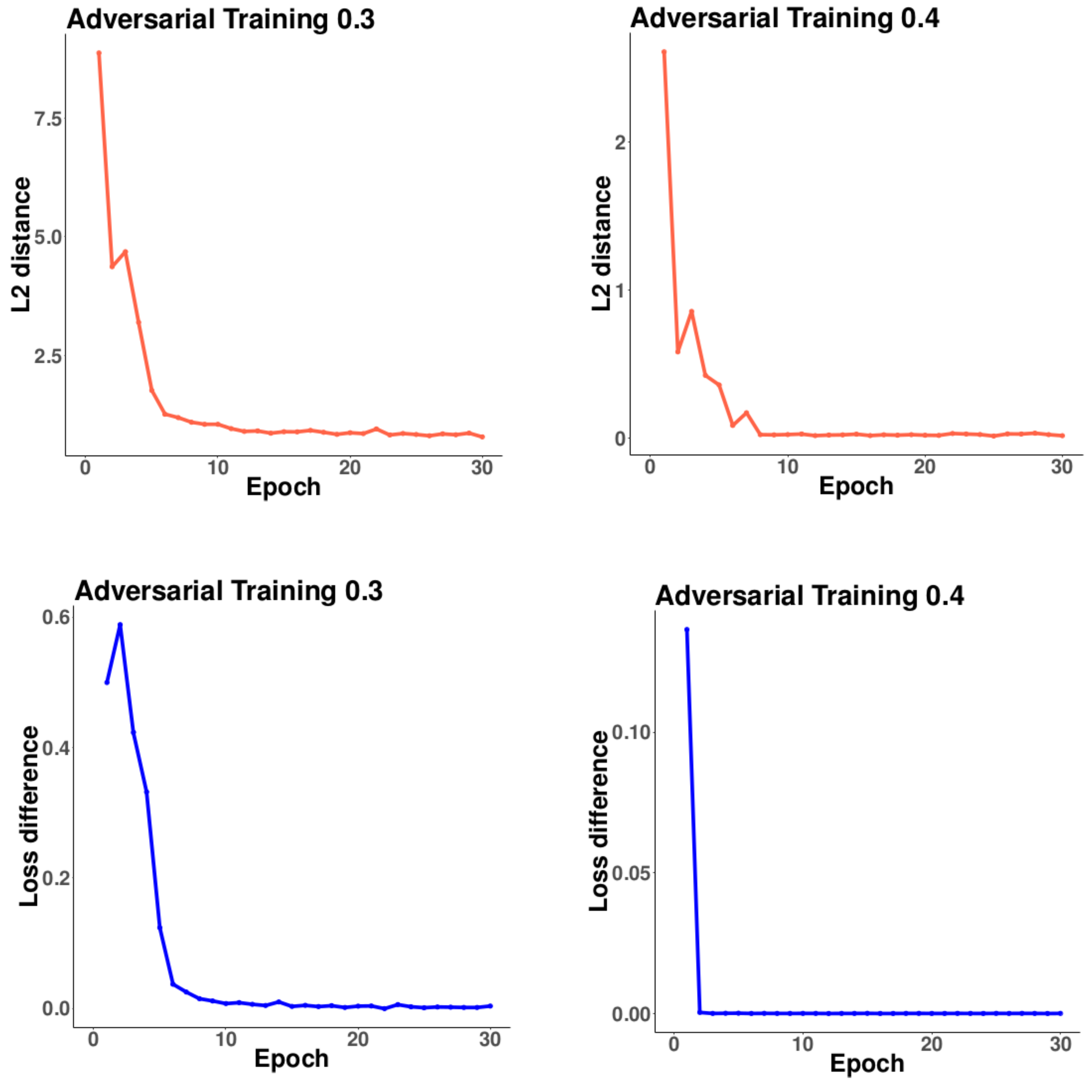}
		\caption{The difference of training adversarial loss and $\ell_2$ distance of weights in two consecutive epochs. These figures empirically indicate the convergence of adversarial training.
		\\
		\\
		\\
		\\
		\\
		\\
		\\
		\\
		\\
		\\
		\\
		\\
		\\
		\\
		\\
		\\
		}
	\end{figure}

	\section{Omitted Figure for the Mode Connectivity}\label{appendix_f}
	
	\begin{figure}[H]
		\centering
		\includegraphics[width=\linewidth]{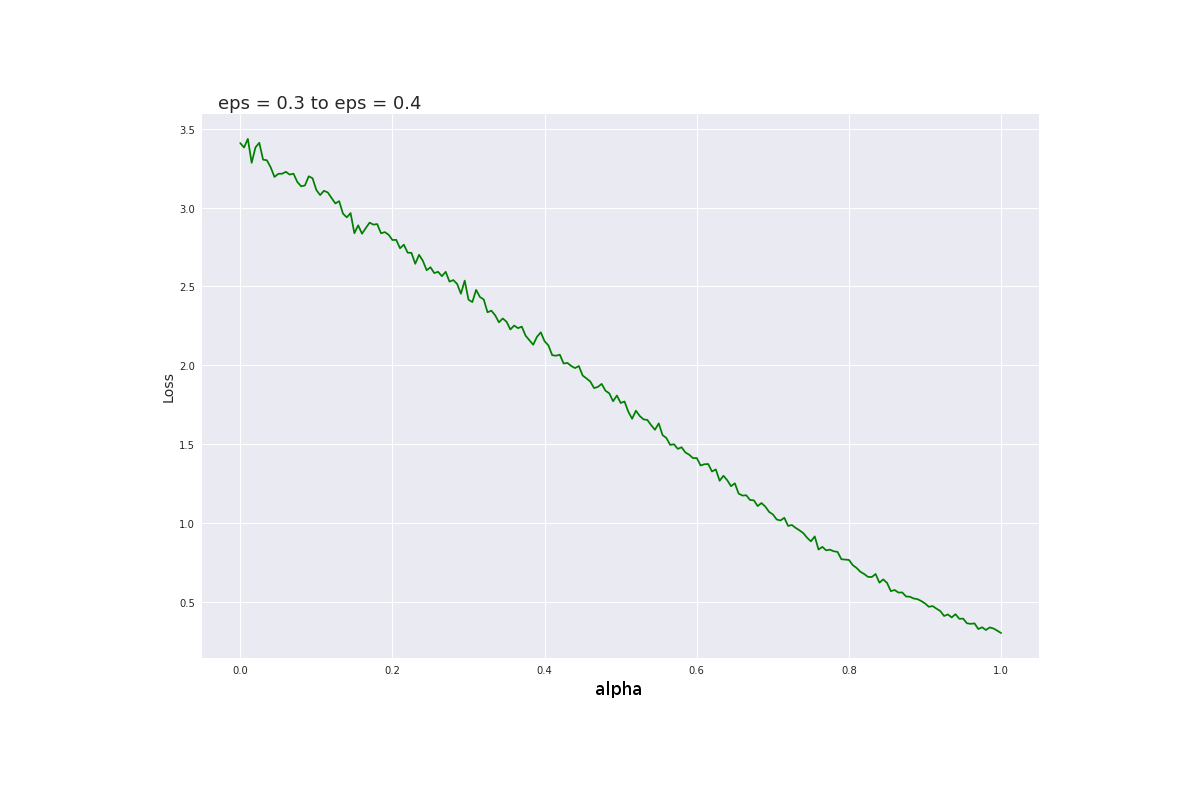}
		\caption{When initialized on weights of an already adversarially trained model with epsilon = $0.3$, $\ell_\epsilon(\alpha)$ could be better approximated by a convex function. Here, for computing the adversarial loss we used the PGD with learning rate = $0.005$ and $200$ steps. This PGD is more accurate than the PGD used for producing a similar figure in the main text. As a result, the function could be better approximated by a convex function.
		\\
		\\
		\\
		\\
		\\
		\\
		\\
		\\
		\\
		\\
		\\
		\\
		\\
		\\
		\\
		\\
		\\
		\\
		\\
		\\
		\\
		\\}
	\end{figure}
	
	\section{Omitted Figure for the Theory Part}\label{appendix_g}
	
	\begin{figure}[H] \label{FIG}
		\centering
		\includegraphics[width=0.8\linewidth]{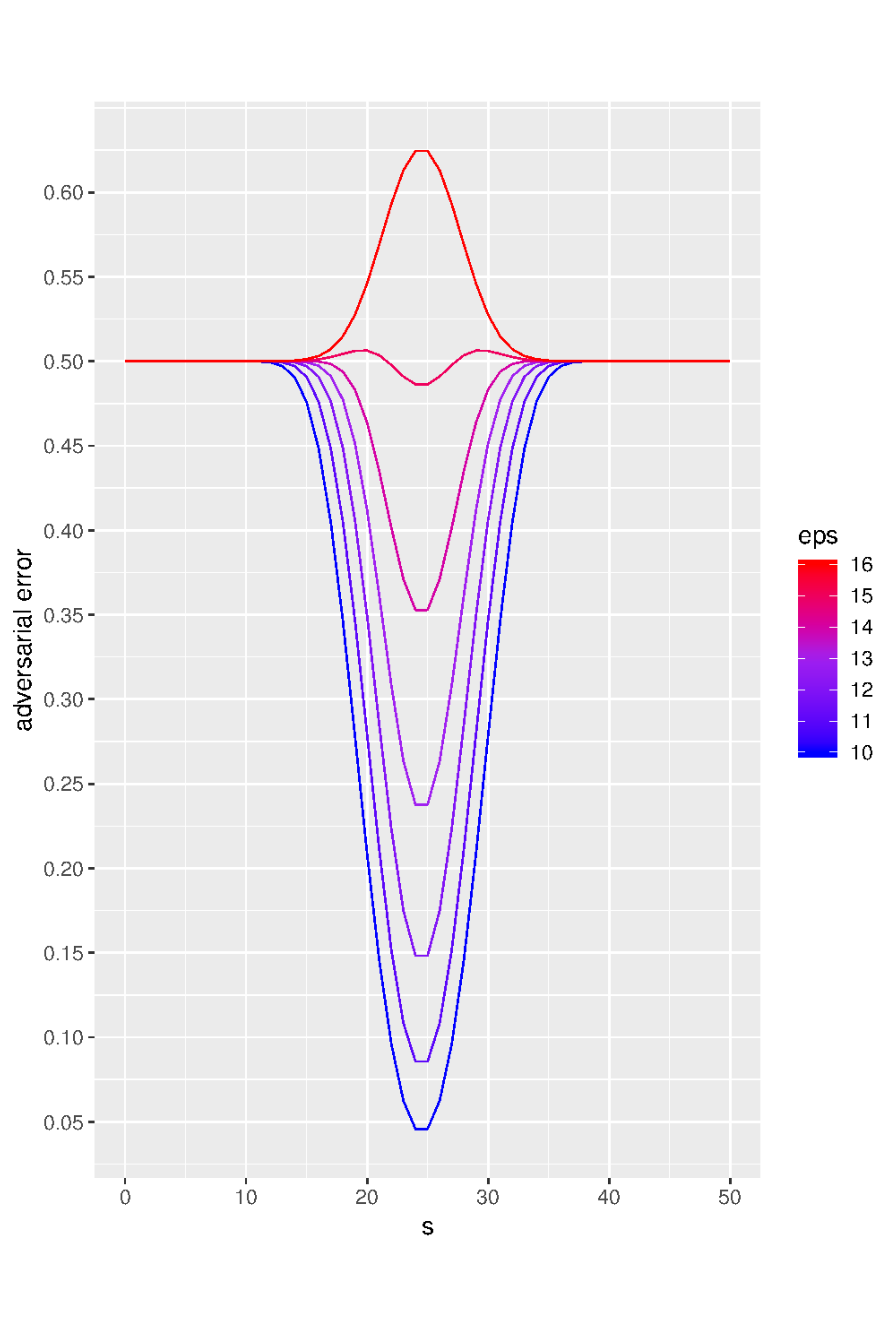}
		\vspace{-40pt}
		\caption{Evaluating the final formula of $\ell_0$ on the multi-dimensional Bernoulli distribution against different hyper-parameter values for $k = 50$ and $t = 0.30$.
		\\
		\\
		\\
		\\}
	\end{figure}

	\section{Omitted Figures for Interpretability}\label{appendix_h}
	
	\begin{figure}[H]
		\centering
		\includegraphics[width=0.9\linewidth]{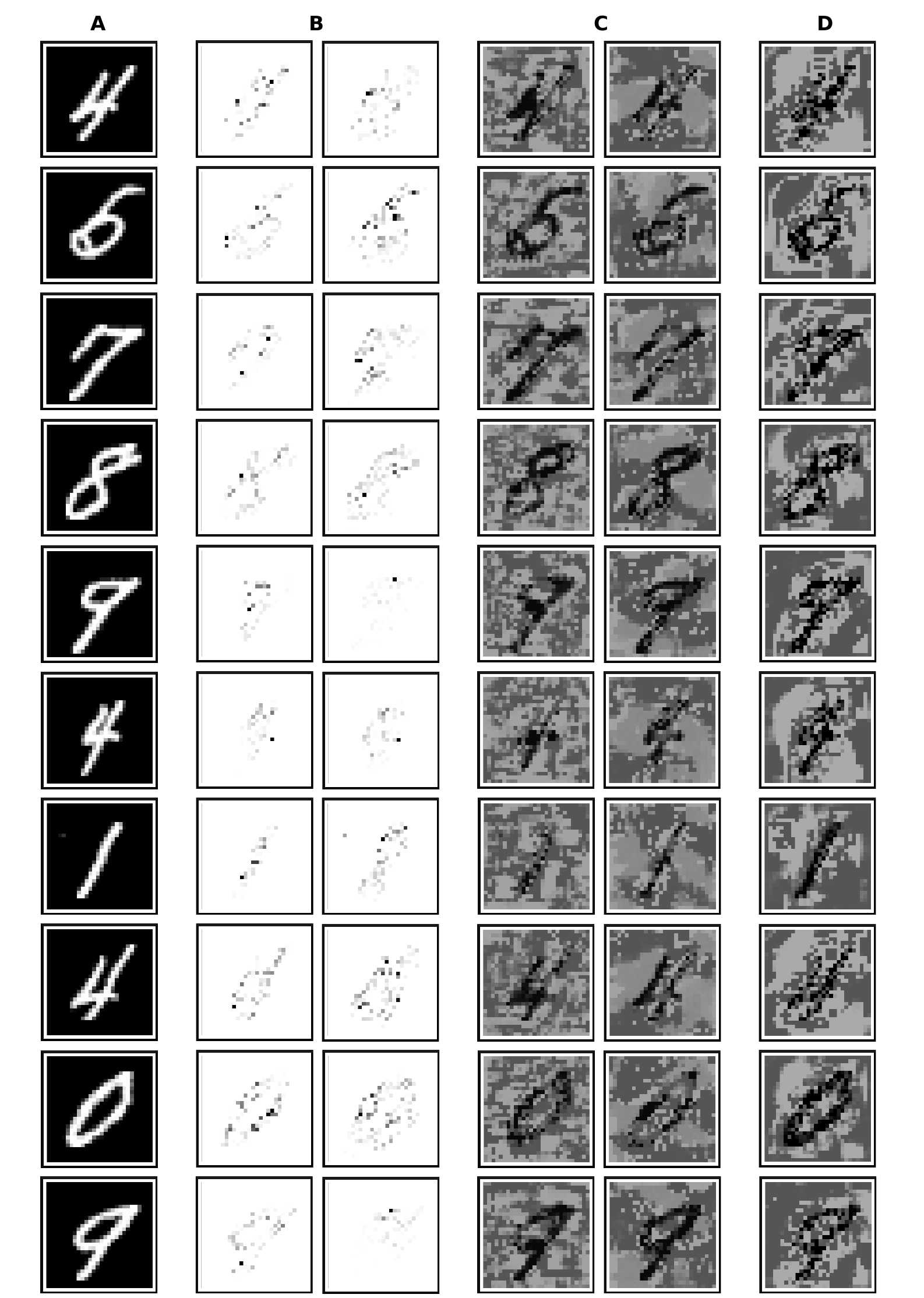}
	\end{figure}
	
	\begin{figure}[H]
		\centering
		\includegraphics[width=0.9\linewidth]{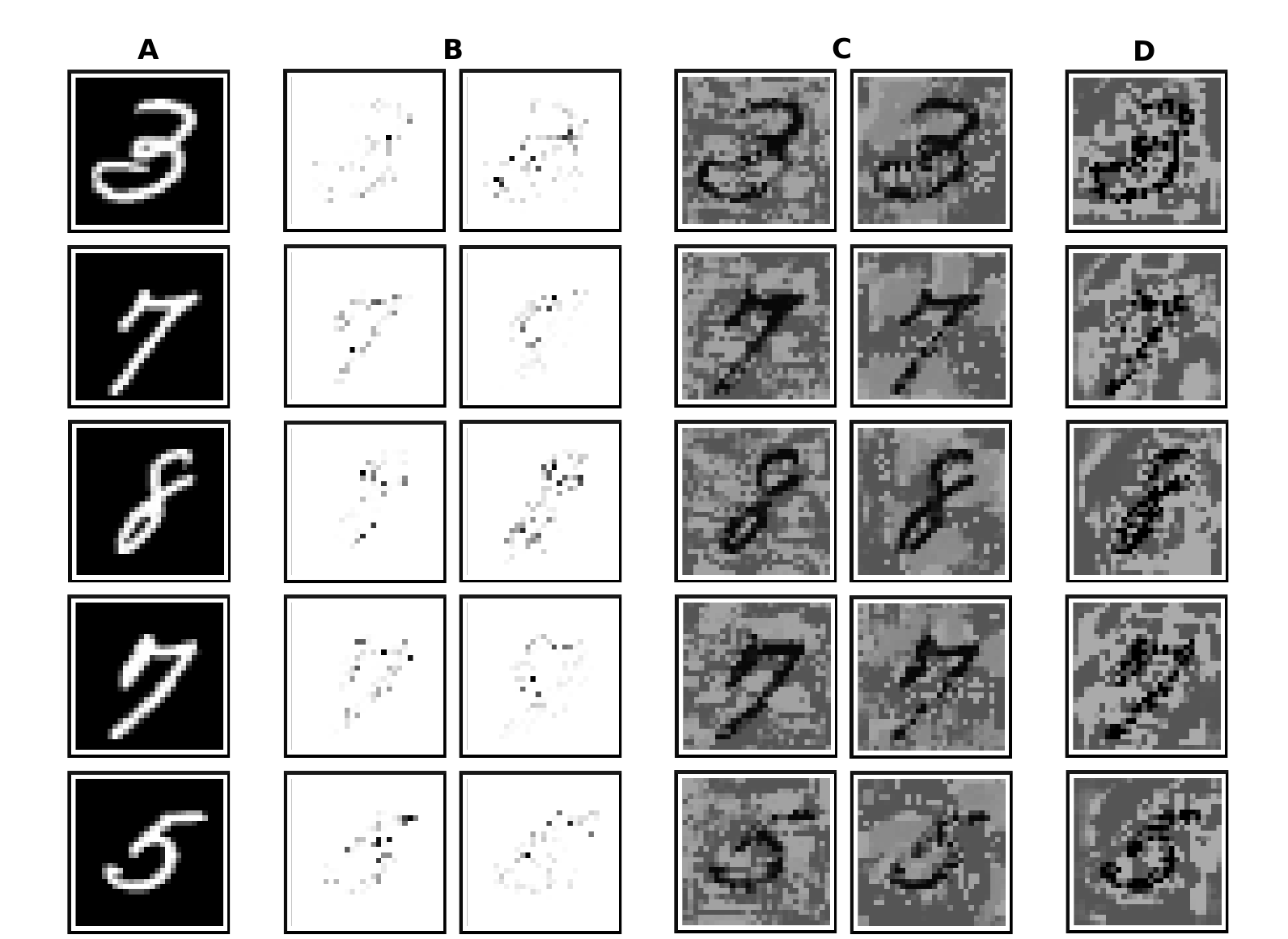}
		\caption{(A): Samples from MNIST. (B) The loss gradient (saliency map) with respect to the perturbed input (of epsilon = 0.3) that is obtained from the model trained on epsilon = $0.3$ (Right) and the model that is trained on epsilon from $0.3$ to $0.4$ (Left). (C) The perturbation (of size epsilon = $0.45$) for the corresponding models in (B). (D) Left: the original image. Right: the perturbations that are obtained from the model trained on epsilon from $0.3$ to $0.5$, based on the attack obtained from the actual gradient that is mentioned in Sec. \ref{carlini}. Overall, training on larger perturbations leads to improved interpretability of both the model and the attacks.}
	\end{figure}

\end{document}